\documentclass[a4paper,fleqn,]{cas-dc}

\usepackage[numbers,sort&compress]{natbib}

\usepackage{amsmath}
\usepackage{bm}
\usepackage{multirow}
\usepackage{colortbl}
\usepackage{rotating}
\usepackage{enumitem}
\usepackage{booktabs}
\usepackage{xcolor}
\usepackage{soul}
\usepackage{hyperref}
\usepackage{scalerel}
\usepackage{tikz}
\usepackage{orcidlink}

\newcommand*\crotate[2]{\rotatebox[origin=c]{90}{\parbox[c][1 pt]{#1 cm}{\centering #2}}}

\begin{document}
\let\WriteBookmarks\relax
\def\floatpagepagefraction{1}
\def\textpagefraction{.001}

\shorttitle{Autonomous UAV Navigation using RL: A Systematic Review}

\shortauthors{AlMahamid \& Grolinger}

\title [mode = title]{Autonomous Unmanned Aerial Vehicle Navigation using Reinforcement Learning: A Systematic Review}        



\author[]{Fadi AlMahamid}[type=editor,
                        auid=000,bioid=1,
                        orcid=0000-0002-6907-7626]
                        
\ead{falmaham@uwo.ca}
\credit{Conceptualization, Methodology, Investigation, Formal Analysis, Validation, Writing - Original Draft, Writing - Review \& Editing}

\author[]{Katarina Grolinger}[auid=001,bioid=2,
                              orcid=0000-0003-0062-8212]
\cormark[1]
\ead{kgrolinger@uwo.ca}
\credit{Supervision, Writing - Review \& Editing, Funding Acquisition}
\affiliation[]{addressline={Department of Electrical and Computer Engineering},
    organization={Western University},
    city={London},
    state={Ontario},
    country={Canada}}

\cortext[1]{Corresponding author.}

\nonumnote{\hyperlink{https://doi.org/10.1016/j.engappai.2022.105321}{https://doi.org/10.1016/j.engappai.2022.105321}}
\nonumnote{The article submitted 5 February 2022 to \textit{Engineering Applications of Artificial Intelligence} Journal; Received in revised form 12 July 2022; Accepted 4 August 2022.}
\nonumnote{0952-1976/\copyright\;2022 Elsevier Ltd. All rights reserved.}

\begin{abstract}
    There is an increasing demand for using Unmanned Aerial Vehicle (UAV), known as drones, in different applications such as packages delivery, traffic monitoring, search and rescue operations, and military combat engagements. In all of these applications, the UAV is used to navigate the environment autonomously - without human interaction, perform specific tasks and avoid obstacles. Autonomous UAV navigation is commonly accomplished using Reinforcement Learning (RL), where agents act as experts in a domain to navigate the environment while avoiding obstacles. Understanding the navigation environment and algorithmic limitations plays an essential role in choosing the appropriate RL algorithm to solve the navigation problem effectively. Consequently, this study first identifies the main UAV navigation tasks and discusses navigation frameworks and simulation software. Next, RL algorithms are classified and discussed based on the environment, algorithm characteristics, abilities, and applications in different UAV navigation problems, which will help the practitioners and researchers select the appropriate RL algorithms for their UAV navigation use cases. Moreover, identified gaps and opportunities will drive UAV navigation research.
\end{abstract}



\maketitle

\begin{table}[!t]
\centering
\begin{tabular}{>{\raggedright}p{0.2\linewidth}|>{\raggedright}p{0.7\linewidth}}
\hline 
\rowcolor[HTML]{EFEFEF} \textbf{Symbol} & \textbf{Description}\tabularnewline
\hline 
\hline 
$\bm{s\in S}$ & State $\bm{s}$ belongs to all possible states $\bm{S}$\tabularnewline
\hline 
$\bm{a\in A}$ & Action $\bm{a}$ belongs to the set of all possible Actions $\bm{A}$\tabularnewline
\hline 
$\bm{r\in R}$ & Reward $\bm{r}$ belongs to the set of all generated Rewards
$\bm{R}$\tabularnewline
\hline 
$\bm{\gamma}$ & Discounted factor $\bm{\gamma}$ decreases the contribution of the future
rewards, where $\bm{0<\gamma<1}$\tabularnewline
\hline 
$\bm{G_{t}}$ & The Expected Summation of the Discounted Rewards; $\bm{G_{t}=\sum_{k=0}^{\infty}\gamma^{k}R_{t+k+1}}$\tabularnewline
\hline 
$\bm{P(s^{\prime},r|s,a)}$ & The probability of the transition to state $\bm{s^{\prime}}$ with
reward $\bm{r}$ from taking action $\bm{a}$ in state $\bm{s}$ at
time $\bm{t}$\tabularnewline
\hline 
$\mbox{\large\ensuremath{\bm{\tau}}}$ & A trajectory $\bm{\tau}$ consists of a sequence of actions and states
pairs, where the actions influence the states, also called an episode.
Each trajectory has a start state and ends in a final state that terminates
the trajectory\tabularnewline
\hline 
$\bm{Q(s,a)}$ & Action-value function expresses the expected return of the state-action pairs $\bm{(s,a)}$; $\bm{Q^{w}(s,a)}$ is $\bm{Q(s,a)}$ parameterized by $\bm{w}$\tabularnewline
\hline 
\end{tabular}
\end{table}

\begin{table}
\begin{tabular}{>{\raggedright}p{0.2\linewidth}|>{\raggedright}p{0.7\linewidth}}
\hline 
\rowcolor[HTML]{EFEFEF} \textbf{Symbol} & \textbf{Description}\tabularnewline
\hline 
\hline
$\bm{V(s)}$ & State-value function is similar to $\bm{Q(s,a)}$ except it measures
how good to be in a state $\bm{s}$; $\bm{V^{w}(s)}$ is a State-value
function parameterized by $\bm{w}$\tabularnewline
\hline 
$\bm{A(s,a)}$ & Advantage-Value function $\bm{A(s,a)}$ measures how good an action is in comparison to alternative actions at a given state; $\bm{A(s,a)=Q(s,a)-V(s)}$\tabularnewline
\hline
$\bm{\pi(a|s)}$ & Stochastic Policy $\bm{\pi}$ is a function that maps the probability
of selecting an action $\bm{a}$ from the state $\bm{s}$. It describes
agent behavior\tabularnewline
\hline 
$\bm{\mu(s)}$ & Deterministic Policy $\bm{\mu}$ is similar to Stochastic Policy $\pi$,
except $\bm{\mu}$ symbol is used to distinguish it from Stochastic
Policy $\bm{\pi}$\tabularnewline
\hline 
$\bm{Q_{\pi}(s,a)}$ & Action-value function $\bm{Q(s,a)}$, when following a policy $\bm{\pi}$\tabularnewline
\hline 
$\bm{V_{\pi}(s)}$ & State-value function $\bm{V(s)}$, when following a policy $\bm{\pi}$\tabularnewline
\hline 
$\mbox{\large\ensuremath{\bm{\rho}}}$ & State visitation probability\tabularnewline
\hline 
$\bm{\sim}$ & Sampled from. For example, $\bm{s\sim\rho}$ means $\bm{s}$ sampled
from state visitation probability $\bm{\rho}$\tabularnewline
\hline 
$\bm{\eta(\pi)}$ & The expected discounted reward following a policy $\bm{\pi}$, similar
to $\bm{G_{t}}$\tabularnewline
\hline 
\end{tabular}
\end{table}

\section{Introduction} \label{sec:introduction}
Autonomous Systems (AS) are systems that can perform desired tasks without human interference, such as robots performing tasks without human involvement, self-driving cars, and delivery drones. AS are invading different domains to make operations more efficient and reduce the cost and risk incurred from the human factor.

An Unmanned Aerial Vehicle (UAV) is an aircraft without a human pilot, mainly known as a drone. Autonomous UAVs have been receiving an increasing interest due to their diverse applications, such as delivering packages to customers, responding to traffic collisions to attain injured with medical needs, tracking military targets, assisting with search and rescue operations, and many other applications.

Typically, UAVs are equipped with cameras, among other sensors, that collect information from the surrounding environment, enabling UAVs to navigate that environment autonomously. UAV navigation training is typically conducted in a virtual 3D environment because UAVs have limited computation resources and power supply, and replacing UAV parts due to crashes can be expensive. 

\begin{figure}[!b]
    \centering
    \includegraphics[width=.8\linewidth]{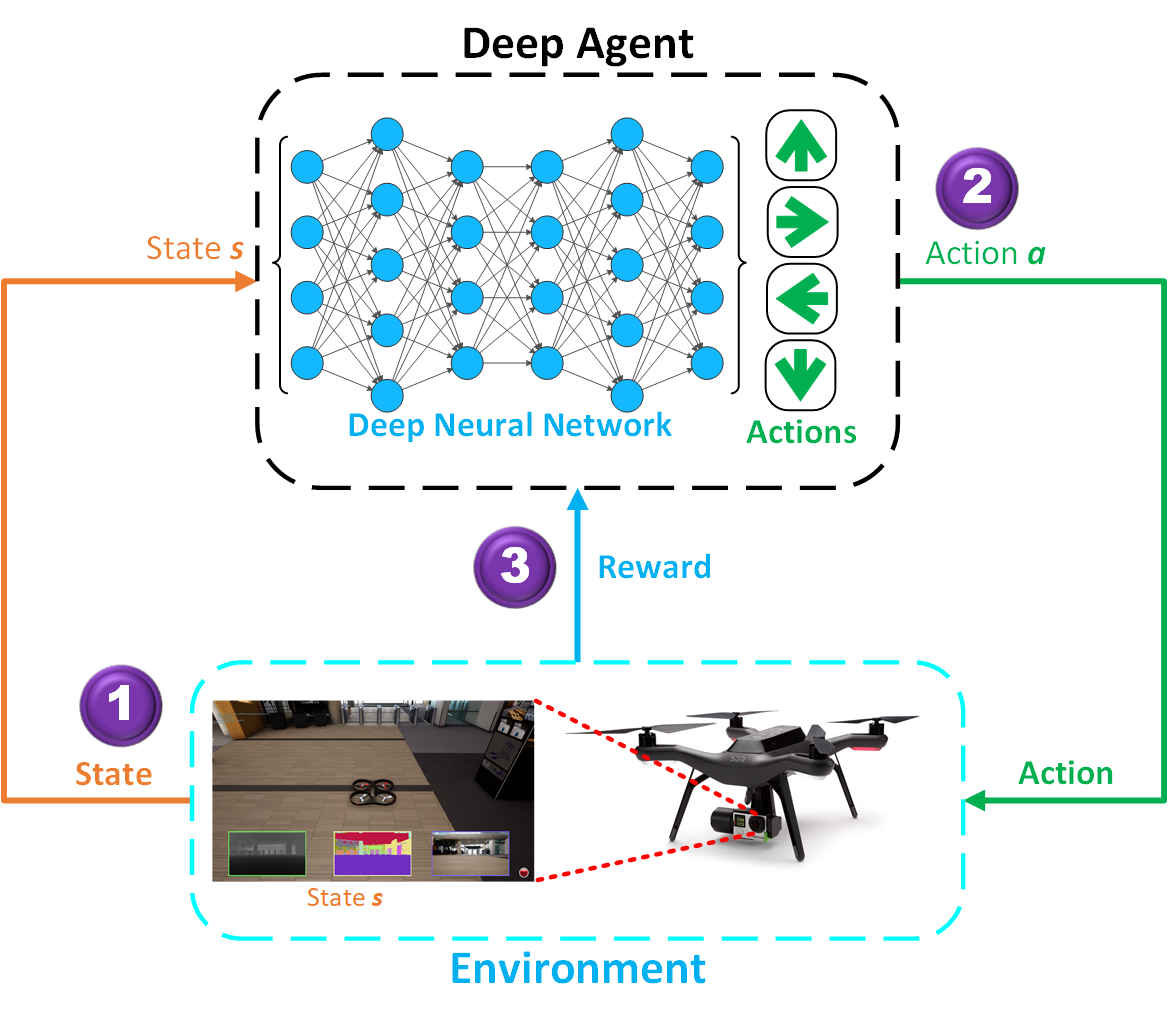}
    \caption{UAV training using deep reinforcement agent}
    \label{fig:uav-drl}
\end{figure}

Different Reinforcement Learning (RL) algorithms are used to train UAVs to navigate the environment autonomously. RL can solve various problems where the agent acts as a human expert in the domain. The agent interacts with the environment by processing the environment's state, responding with an action, and receiving a reward. UAV cameras and sensors capture information from the environment for state representation. The agent processes the captured state and outputs an action that determines the UAV movement's direction or controls the propellers' thrust, as illustrated in Figure \ref{fig:uav-drl}.

The research community provided a review of different UAV navigation problems, such as Visual UAV navigation \citep{lu2018survey, zeng2020survey}, UAV Flocking \citep{azoulay2021machine} and Path Planning \citep{aggarwal2020path}. Nevertheless, to the best of the authors' knowledge there is no survey related to applications of RL in UAV navigation. Hence, this paper aims to provide a comprehensive and systematic review on the application of various RL algorithms to different autonomous UAV navigation problems. This survey has the following contributions:

\begin{itemize}[noitemsep,nolistsep]
    \item Help the practitioners and researchers to select the right algorithm to solve the problem on hand based on the application area and environment type.
    \item Explain primary principles and characteristics of various RL algorithms, identify relationships among them, and classify them according to the environment type.
    \item Discuss and classify different RL UAV navigation frameworks according to the problem domain.
    \item Recognize the various techniques used to solve different UAV autonomous navigation problems and the different simulation tools used to perform UAV navigation tasks.
\end{itemize}

The remainder of the paper is organized as follows: Section \ref{sec:review-process}  presents the systematic review process, Section \ref{sec:background} introduces RL, Section \ref{sec:uav-navigation} provides a comprehensive review of the application of various RL algorithms and techniques in autonomous UAV navigation, Section \ref{sec:frameworks-sumi-soft} discusses the UAV Navigation Frameworks and simulation software, Section \ref{sec:rl-algorithms} classifies RL algorithm and discusses the most prominent algorithms, Section \ref{sec:alg-selection} explains RL algorithms selection process, and Section \ref{sec:challanges-opportunities} identifies challenges and research opportunities. Finally, Section \ref{sec:conclusion} concludes the paper.

\section{Review Process} \label{sec:review-process}
This section described the inclusion criteria, paper identification process, and threats to validity.
\subsection{Inclusion Criteria and Identification of Papers}
The study's main objective is to analyze the application of Reinforcement Learning in UAV navigation and provide insights into RL algorithms. Therefore, the survey considered all papers in the past five years (2016-2021) written in the English language that include the following terms combined, alongside with their variations: \textit{Reinforcement Learning}, \textit{Navigation}, and \textit{UAV}.

In contrast, RL algorithms are listed based on the authors' domain knowledge of the most prominent algorithms and by going through the related work of the identified algorithms with no restriction to the publication time to include a large number of algorithms.

The identification process of the papers went through the following stages:
\begin{itemize}[noitemsep,nolistsep]
    \item First stage: The authors identified all studies that strictly applied RL to UAV Navigation and acknowledged that model-free RL is typically utilized to tackle UAV navigation challenges, except for a single article \citep{lou2016adaptive} that employs model-based RL. Therefore, the authors choose to concentrate on model-free RL and exclude research irrelevant to UAV Navigation, such as UAV networks and traditional optimization tools and techniques \citep{guerra2021networks, guerra2021real, guerra2020dynamic, liu2020distributed, zhang2020self}.
    \item Second stage: The authors listed all RL algorithms based on authors' knowledge of the most prominent algorithms, the references of recognized algorithms, then identified the corresponding paper of each algorithm.
    \item Third stage: the authors identified how RL is used to solve different UAV navigation problems, classified the work, and then recognized more related papers using exiting work references.   
\end{itemize}

IEEE Xplore and Scopus were the primary sources of papers' identification between 2016 and 2021. The search query was applied using different terminologies that are used to describe the UAV alternatively, such as \textit{UNMANNED AERIAL VEHICLE}, \textit{DRONE}, \textit{QUADCOPTER}, or \textit{QUADROTOR}, and these terms are cross-checked with \textit{REINFORCEMENT LEARNING}, and \textit{NAVIGATION}, which resulted in a total of 104 papers. After removing 15 duplicate papers and 5 unrelated papers, the count became 84.

The authors identified another 75 papers that mainly describe the RL algorithms based on the authors' experience and the references list of the recognized work, using Google Scholar as the primary search engine. While RL for UAV navigation studies were restricted to five years, all RL algorithms are included as many are still extensively used regardless of their age. The search was completed in November 2021, with a total of 159 papers after all exclusions.

\subsection{Threats to Validity}
Despite the authors' effort to include all relevant papers, the study might be subject to the following main threats to the validity:

\begin{itemize}[noitemsep,nolistsep]
    \item Location bias: The search for papers was performed using two primary digital libraries (databases), IEEE Xplore and Scopus, which might limit the retrieved papers based on the published journals, conferences, and workshops in the database.
    \item Language bias: Only papers published in English are included.
    \item Time Bias: The search query is only limited to retrieving papers between 2016 and 2021, which results in excluding relevant papers published before 2016.
    \item Knowledge reporting bias: The research papers of RL algorithms are identified using authors' knowledge of variant algorithms and the related work in the recognized algorithms. It is hard to pinpoint all algorithms utilizing a search query, which could result in missing some RL algorithms.
\end{itemize}

\section{Reinforcement Learning} \label{sec:background}
RL can be explained using the Markov Decision Process (MDP), where a RL agent learns through experience by taking actions in the environment, causing a change in the environment's state, and receiving a reward for the action taken to measure the success or failure of the action. Equation \ref{equ:mdp} defines the transition probability from state $\bm{s}$ by taking the action $\bm{a}$ to the new state $\mathbf{s^{\prime}}$ with a reward $\bm{r}$, for all $s^{\prime} \in S, \; s \in S, \; r \in R, \; a \in A(s)$ \citep{almahamid2021reinforcement}.

\begin{equation}
    \label{equ:mdp}
    P(s^{\prime},r|s,a) = Pr\{S_t = s^{\prime}, R_t=r| S_{t-1}=s, A_{t-1}=a\}    
\end{equation}

The reward $\bm{R}$ is generated using a reward function, which can be expressed as a function of the action $R(a)$, or as a function of action-state pairs $R(a,s)$. The reward helps the agent learn good actions from bad actions, and as the agent accumulates experience, it starts taking more successful actions and avoiding bad ones \citep{almahamid2021reinforcement}.

All actions the agent takes from a start state to a final (terminal) state make an \textit{episode} (trajectory). The goal of MDP is to maximize the expected summation of the discounted rewards by adding all the rewards generated from an episode. However, sometimes the environment has an infinite horizon, where the actions cannot be divided into episodes. Therefore, using a discounted factor (multiplier) $\bm{\gamma}$ to the power $\bm{k}$, where $\bm{\gamma \in [0,1]}$ as expressed in Equation \ref{equ:sum-disc-reward} helps the agent to emphasize the reward at the current time step and reduce the reward value granted at future time steps, and, moreover, helps the expected summation of discounted rewards to converge if the horizon is infinite \citep{almahamid2021reinforcement}.

\begin{equation}
    \label{equ:sum-disc-reward}
    G_t = E \left[\sum_{k=0}^{\infty} \gamma^{k}  R_{t+k+1}\right]
\end{equation}

The following subsections introduce important reinforcement learning concepts.

\subsection{Policy and Value Function}
A policy $\bm{\pi}$ defines the agent's behavior by defining the probability of taking action $\bm{a}$ while being in a state $\bm{s}$, which is expressed as $\bm{\pi(a|s)}$. The agent evaluates its behavior (action) using a value function, which can be either \textit{state-value function}, which estimates how good it is to be in state $\bm{s}$ after executing an action $\bm{a}$, or using a \textit{action-value function} that measures how good it is to select action $\bm{a}$ while being in a state $\bm{s}$. The value produced by the action-value function in Equation \ref{equ:action-value-func} is known as the Q-value and is expressed in terms of the expected summation of the discounted rewards \citep{almahamid2021reinforcement}.
 
\begin{equation}
    \label{equ:action-value-func}
    Q_\pi(s,a) = E_\pi \left[\sum_{k=0}^{\infty} \gamma^{k}  R_{t+k+1} \; | \; S_t = s, A_t = a \right]
\end{equation}

Since the objective is to maximize the expected summation of discounted rewards under the optimal policy $\bm{\pi}$, the agent tries to find the optimal Q-value $\bm{Q_*(s,a)}$ as defined in Equation \ref{equ:optimal-policy}. This optimal  Q-value must satisfy the \textit{Bellman Optimality Equation} \ref{equ:bellman-optimality} which is defined as the sum of the expected reward received from executing the current action $\bm{R_{t+1}}$, and sum of all future rewards (discounted) received from any possible future state-action pairs $\bm{(s^\prime,a^\prime)}$. In other words, the agent tries to select the actions that grant the highest rewards in an episode. In general, selecting the optimal value means selecting the action with the highest Q-value; however, the action with the highest Q-value sometimes might not lead to better rewarding actions in the future \citep{almahamid2021reinforcement}.

\begin{equation}
    \label{equ:optimal-policy}
    Q_*(s,a) = \underset{\pi}{max} \; Q_\pi(s,a)
\end{equation}

\begin{equation}
    \label{equ:bellman-optimality}
    Q_*(s,a) = E \left[ R_{t+1}  + \gamma \; \underset{a^\prime}{max} \; Q_*(s^\prime,a^\prime) \right]  
\end{equation}

\subsection{Exploration vs Exploitation}
Exploration vs. Exploitation may be demonstrated using the multi-armed bandit dilemma, which accurately portrays the behavior of a person experiencing their first slot machine experience. The money (reward) player receives early in the game is unrelated to any previously selected choices, and as the player develops a comprehension of the reward, he/she begins selecting choices that contribute to earning a greater reward. The choices made randomly by the player to acquire knowledge might be defined as the player \textit{Exploring} the environment. In contrast, the player's \textit{Exploiting} the environment is described as the options selected based on his/her experience. 

The RL agent needs to find the right balance between exploration and exploitation to maximize the expected return of rewards. Constantly exploiting the environment and selecting the action with the highest reward does not guarantee that the agent performs the optimal action because the agent may miss out on a higher reward provided by future actions taking alternative sets of actions in the future. Finding the ratio between exploration and exploitation can be defined through different strategies such as $\epsilon$-greedy strategy, Upper Confidence Bound (UCB), and Gradient Bandits \citep{sutton2018reinforcement}.

\subsection{Experience Replay}
RL agent does not need data to learn; rather, it learns from experiences by interacting with the environment. The agent experience $\bm{e}$ can be formulated as tuple $\bm{e(s,a,s^\prime,r)}$, which describes the agent taking an action $\bm{a}$ at a given state $\bm{s}$ and receiving a reward $\bm{r}$ for the performed action and causing a new state $\bm{s^\prime}$. \textit{Experience Replay (ER)} \citep{lin1992self} is a technique that suggests storing experiences in a replay memory (buffer) $\bm{D}$ and using a batch of uniformly sampled experiences for RL agent training. 

On the other hand, \textit{Prioritized Experience Replay (PER)} \citep{schaul2015prioritized} prioritizes experiences according to their significance using \textit{Temporal Difference error (TD-error)} and replays experiences with lower TD-error to repeatedly train the agent, which improves the convergence.  

\subsection{On-Policy vs Off-Policy} \label{sec:on-policy-vs-off-policy}
In order to interact with the environment, the RL agent attempts to learn two policies: the first one is referred to as the target policy $\bm{\theta(a|s)}$, which the agent learns through the value function, and the second one is referred to as the behavior policy $\bm{\beta(a|s)}$, which the agent uses for action selection when interacting with the environment. 

A RL algorithm is referred to as \textit{on-policy algorithm} when the same target policy $\bm{\theta(a|s)}$ is employed to collect the training sample and to determine the expected return. In contrast, \textit{off-policy algorithms} are those where the training sample is collected in accordance to the behavior policy $\bm{\beta(a|s)}$, and the expected reward is generated using the target policy $\bm{\theta(a|s)}$ \citep{silver2014deterministic}. Another main difference is that Off-policy algorithms can reuse past experiences and do not require all the experiences within an episode (full episode) to generate training samples, and the experiences can be collected from different episodes.

\subsection{Deep Reinforcement Learning}
Deep Reinforcement Learning (DRL) uses deep agents to learn the optimal policy where it combines artificial Neural Networks (NN) with Reinforcement Learning (RL). The NN type used in DRL varies from one application to another depending on the problem being solved, inputs type (state), and the number of inputs passed to the NN. For example, the RL framework can be integrated with Convolutional Neural Network (CNN) to process images representing the environment's state or combined with Recurrent Neural Network (RNN) to process inputs over different time steps.

The NN loss function, also known as the Temporal Difference (TD), is generically computed by finding the difference between the output of the NN $\bm{Q(s,a)}$ and the optimal Q-value $\bm{Q_*(s,a)}$ obtained from the Bellman equation as shown in Equation \ref{equ:loss-func} \citep{almahamid2021reinforcement}:

\begin{equation}
    \label{equ:loss-func}
    \overset{Target}{\overbrace{E \left[ R_{t+1}  + \gamma \; \underset{a^\prime}{max} \; Q_*(s^\prime,a^\prime) \right]}} \; - \;  \overset{Predicted}{\overbrace{E \left[\sum_{k=0}^{\infty} \gamma^{k}  R_{t+k+1}\right]}}
\end{equation}

The architecture of the deep agent can be simple or complex based on the problem at hand, where a complex architecture combines multiple NN. But what all deep agents have in common is that they receive the state as an input, then they output the optimal action and maximize the discounted return of rewards.

The application of Deep NN to the RL framework enabled the research community to solve more complex problems in autonomous systems that were hard to solve before and achieve better performance than previous state-of-the-art, such as drone navigation and avoiding obstacles using images received from the drone's monocular camera.

\section{Autonomous UAV Navigation using DRL} \label{sec:uav-navigation}
Different DRL algorithms and techniques were used to solve various problems in autonomous UAV navigation, such as UAV control, obstacle avoidance, path planning, and flocking. The DRL agent acted as an expert in all of these problems, selecting the best action that maximizes the reward to achieve the desired objective. The input and the output of the DRL algorithm are generally determined based on the desired objective and the implemented technique.

\begin{figure*}[!t]
    \centering
    \includegraphics[width=1\linewidth]{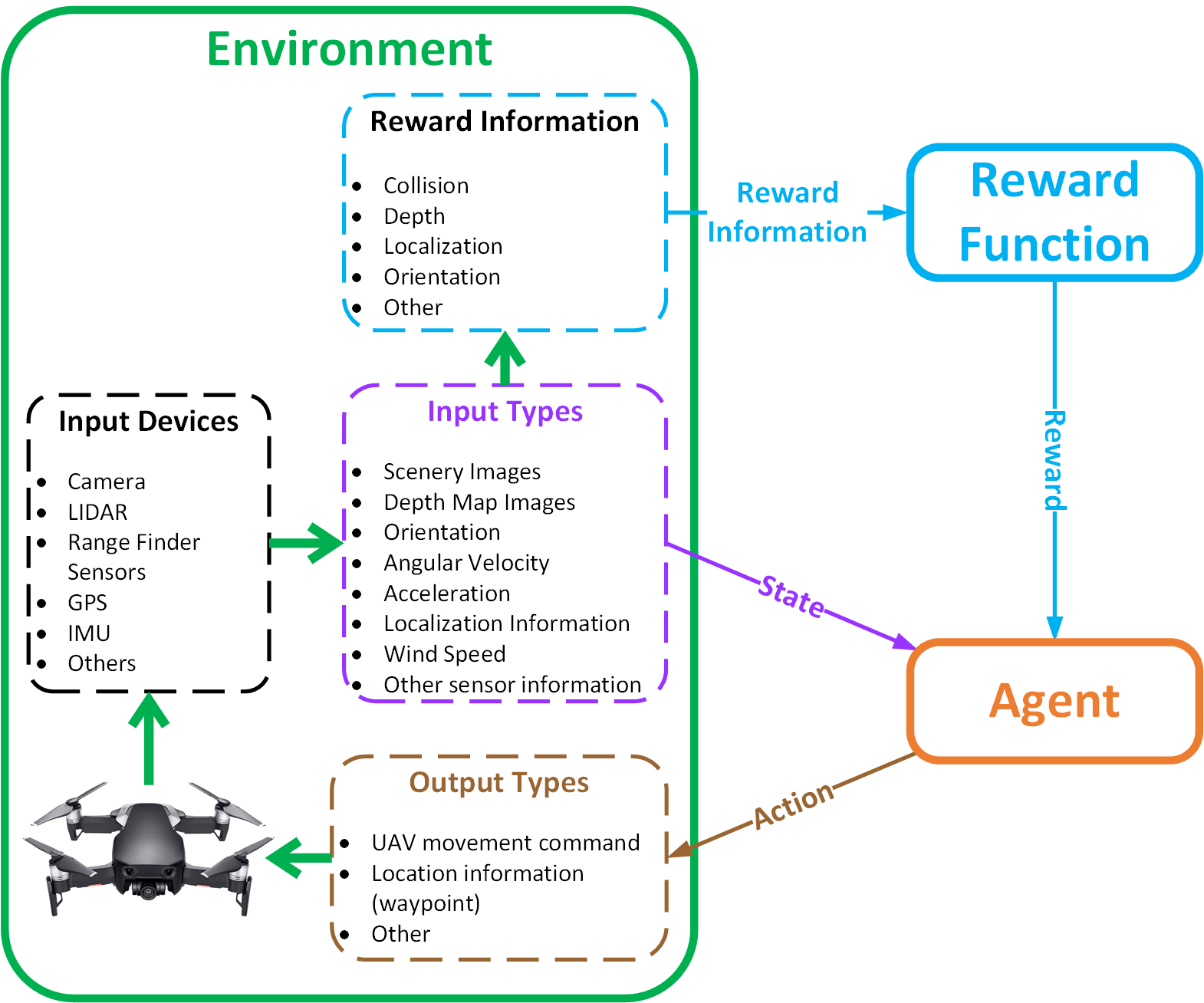}
    \caption{RL agent design for UAV navigation task}
    \label{fig:rl-agent-design}
\end{figure*}

RL agent design for UAV navigation depicted in Figure \ref{fig:rl-agent-design} shows different UAV input devices used to capture the state processed by the RL agent. The agent produces action values that can be either the movement values of the UAV or the waypoint values where the UAV needs to relocate. Once the agent executes the action in the environment, it receives the new state and the generated reward based on the performed action. The reward function is designed to generate the reward subject to the intended objective while using various information from the environment. The agent design ('Agent' box in the figure) is influenced by the RL algorithms discussed in Section \ref{sec:rl-algorithms} where the agent components and inner working varies from one algorithm to another.

Table \ref{tab:rl_application_to_nav} summarizes the application of RL to different UAV navigation tasks (objectives), and the following subsections discuss the UAV navigation tasks in more detail. As seen from this table, most of the research focused on two UAV navigation objectives: 1) Obstacle avoidance using various UAV sensor devices such as cameras and LIDARs and 2) Path planning to find the optimal or shortest route.

\begin{table*}[t]
    \centering
    \caption{DRL application to different UAV Navigation tasks}
    \label{tab:rl_application_to_nav}
    \begin{tabular}{l|l|l}
    \hline 
    \rowcolor[HTML]{EFEFEF} \textbf{Objective} & \textbf{Sub-Objective} & \textbf{Paper}\tabularnewline
    \hline 
    \hline 
    \textbf{UAV Control} & Controlling UAV flying behavior (attitude control) & \cite{zhou2020efficient, karthik2020reinforcement, lou2016adaptive, deshpande2020developmental, camci2019learning, li2019optimal, greatwood2019reinforcement, koch2019reinforcement}\tabularnewline
    \hline 
    \multirow{2}{*}{\textbf{Obstacle Avoidance}} & Obstacle avoidance using images and sensor information & \cite{salvatore2020neuro,bouhamed2020uav,huang2019autonomous,shin2019automatic,wang2017autonomous,wang2019autonomous,anwar2020autonomous,bouhamed2020autonomous,yang2020autonomous,li2019autonomous,chen2020collision,grando2020deep,camci2020deep,wang2020deep,cetin2019drone,morad2021embodied,yan2020flocking,yoon2019hierarchical,williams2017information,he2020integrated,singla2019memory,wu2018navigating,anwar2018navren,zhou2018neural,yijing2017q,villanueva2019deep,walvekar2019vision,zhou2019vision,hasanzade2021dynamically, munoz2019deep, hodge2021deep, doukhi2021deep, bakale2020indoor, maxey2019navigation, zhao2021reinforcement, greatwood2019reinforcement, tong2021uav}\tabularnewline
    \cline{2-3} \cline{3-3} 
     & Obstacle avoidance while considering the battery level & \cite{bouhamed2020ddpg}\tabularnewline
    \hline 
    \multirow{4}{*}{\textbf{Path Planning}} & Local and global path planning (finding the shortest/optimal route) & \cite{walker2019deep, bouhamed2020generic, bouhamed2020uav, shin2019automatic, zhang2020iadrl, yu2019navigation, li2018path, wu2018navigating, sacharny2019optimal, camci2019planning, guerra2020reinforcement, cui2021uav, hasanzade2021dynamically, wang2021pretrained, eslamiat2019autonomous, bakale2020indoor, tong2021uav}\tabularnewline
    \cline{2-3} \cline{3-3} 
     & Path planning while considering the battery level & \cite{bouhamed2020uav, imanberdiyev2016autonomous, abedin2020data}\tabularnewline
    \cline{2-3} \cline{3-3} 
     & Find fixed or moving targets (points of interest) & \cite{andrew2018deep, pham2018reinforcement, guerra2020reinforcement, kulkarni2020uav, peake2020wilderness, akhloufi2019drones, tong2021uav}\tabularnewline
    \cline{2-3} \cline{3-3} 
     & Landing the UAV on a selected point & \cite{polvara2018toward,polvara2019autonomous,lee2018vision}\tabularnewline
    \hline 
    \multirow{6}{*}{\textbf{Flocking}} & Maintain speed and orientation with other UAVs (formation) & \cite{wang2018deep, lee2020autonomous, yan2020flocking, madridano2021software}\tabularnewline
    \cline{2-3} \cline{3-3} 
     & Obstacle avoidance & \cite{wang2020two, madridano2021software}\tabularnewline
    \cline{2-3} \cline{3-3} 
     & Target tracking & \cite{moon2021deep, liu2019distributed, omi2021introduction, viseras2021wildfire, bonnet2019uav}\tabularnewline
    \cline{2-3} \cline{3-3} 
     & Flocking while considering the battery level & \cite{liu2019distributed}\tabularnewline
    \cline{2-3} \cline{3-3} 
     & Covering geographical region & \cite{liu2019distributed, fan2020prioritized}\tabularnewline
    \cline{2-3} \cline{3-3} 
     & Path planning and finding the safest route & \cite{majd2018integrating, madridano2021software}\tabularnewline
    \hline 
    \end{tabular}
\end{table*}

\subsection{UAV Control}
RL is used to control the movement of the UAV in the environment by applying changes to the flight mechanics of the UAV, which varies based on the UAV type. In general UAVs can be classified based on the flight mechanics into 1) Multirotor, 2) Fixed-Wing, and 3) single-rotor, and 4) fixed-wing hybrid Vertical Take-Off and Landing (VTOL) \citep{chapman2016dronetypes}.

\begin{figure}[!t]
    \centering
    \includegraphics[width=.9\linewidth]{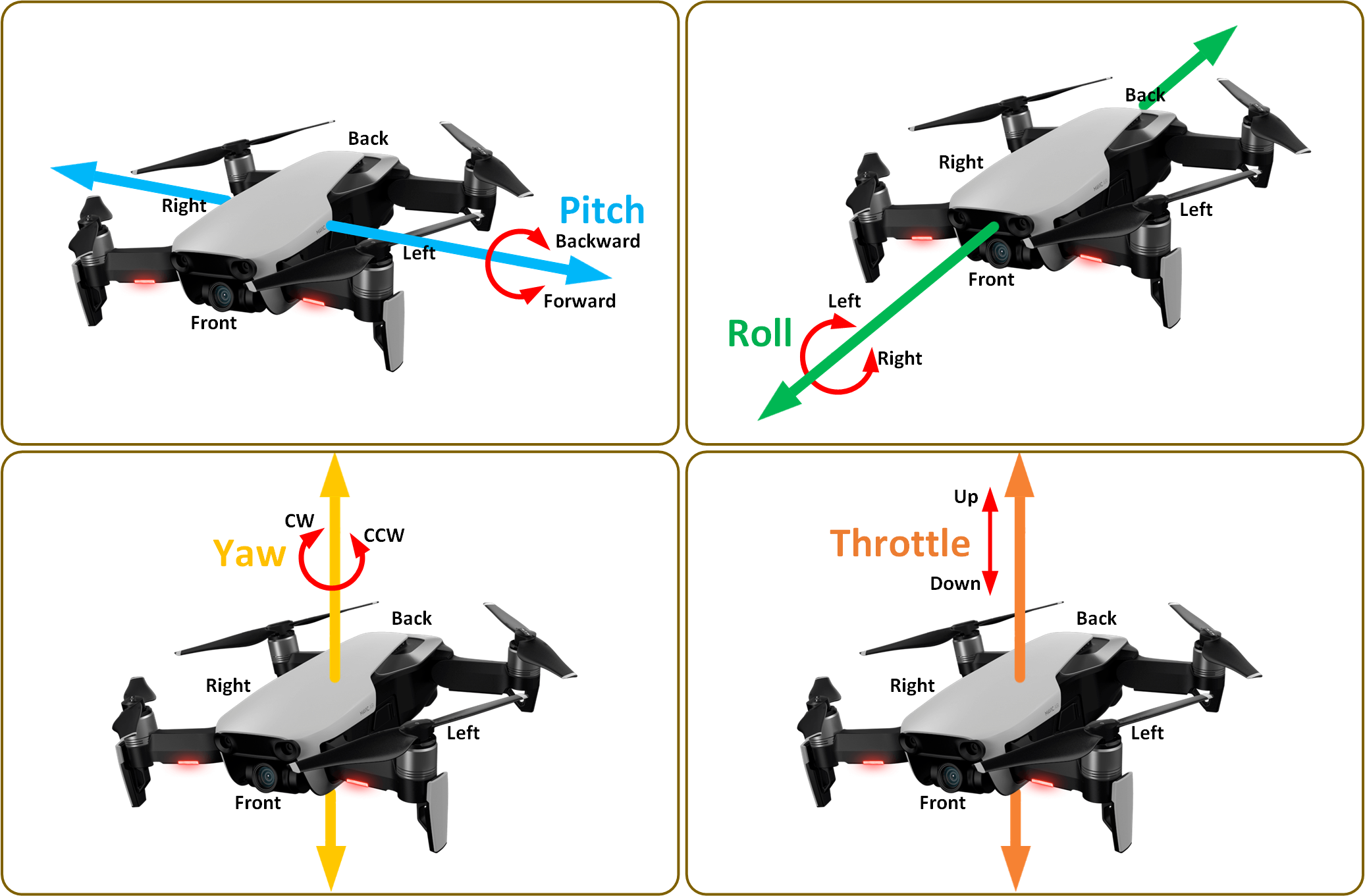}
    \caption{Multirotor Flight Mechanics}
    \label{fig:multirotor-flight-mechanics}
\end{figure}

Multirotor, also known as multicopter or drone, uses more than two rotors to control the flight mechanics by applying different amounts of thrust to the rotors causing changes in principal axes leading to four UAV movements 1) pitch, 2) roll, 3) yaw, and 4) throttle as explained in Figure \ref{fig:multirotor-flight-mechanics}. Similarly, single-rotor and fixed-wing hybrid VTOL apply changes to different rotors to generate the desired movement, except they both use tilt-rotor(s) and wings in fixed-wing hybrid VTOL. On the other hand, fixed-wing can only achieve three actions pitch, roll, and yaw, where they take off by generating enough speed that causes the air-dynamics to lift-up the UAV.

Quad-rotors has four propellers: two diagonal propellers rotate clockwise and the other two propellers rotate counter-clockwise causing the throttle action. When the propellers generate a thrust more significant than the UAV weight they cause elevation, and when the thrust power equals the UAV weight, the UAV stops elevation and starts hovering in place. In contrast, if all propellers rotate in the same direction, they cause a yaw action in the opposite direction, as shown in Figure \ref{fig:yaw-throttle-mechanics}. 

The steps described in Figure \ref{fig:uav-control} depicts the RL process used to control the UAV, which depends on the used RL algorithm, but the most important takeaway is that RL uses the UAV state to produce actions. These actions are responsible for moving the UAV in the environment and can be either direct changes in the value of pitch, roll, yaw, and throttle values or indirect changes that require transformation to commands understood by the UAV.

\subsection{Obstacle Avoidance}
Avoiding obstacles is an essential task required by the UAV to navigate any environment, which can be achieved by estimating the distance to the objects in the environment using different devices such as front-facing cameras or distance sensors. The output generated by these different devices provides input to the RL algorithm and plays a significant role in the NN architecture.

Lu \textit{et al.} \citep{lu2018survey} described different front-facing cameras such as monocular cameras, stereo cameras, and RGB-D cameras that a UAV can use. Each camera type produces a different image type used as raw input to the RL agent. However, regardless of the camera type, these images can be preprocessed using computer vision to produce specific image types as described below:

\begin{figure}[!t]
    \centering
    \includegraphics[width=.9\linewidth]{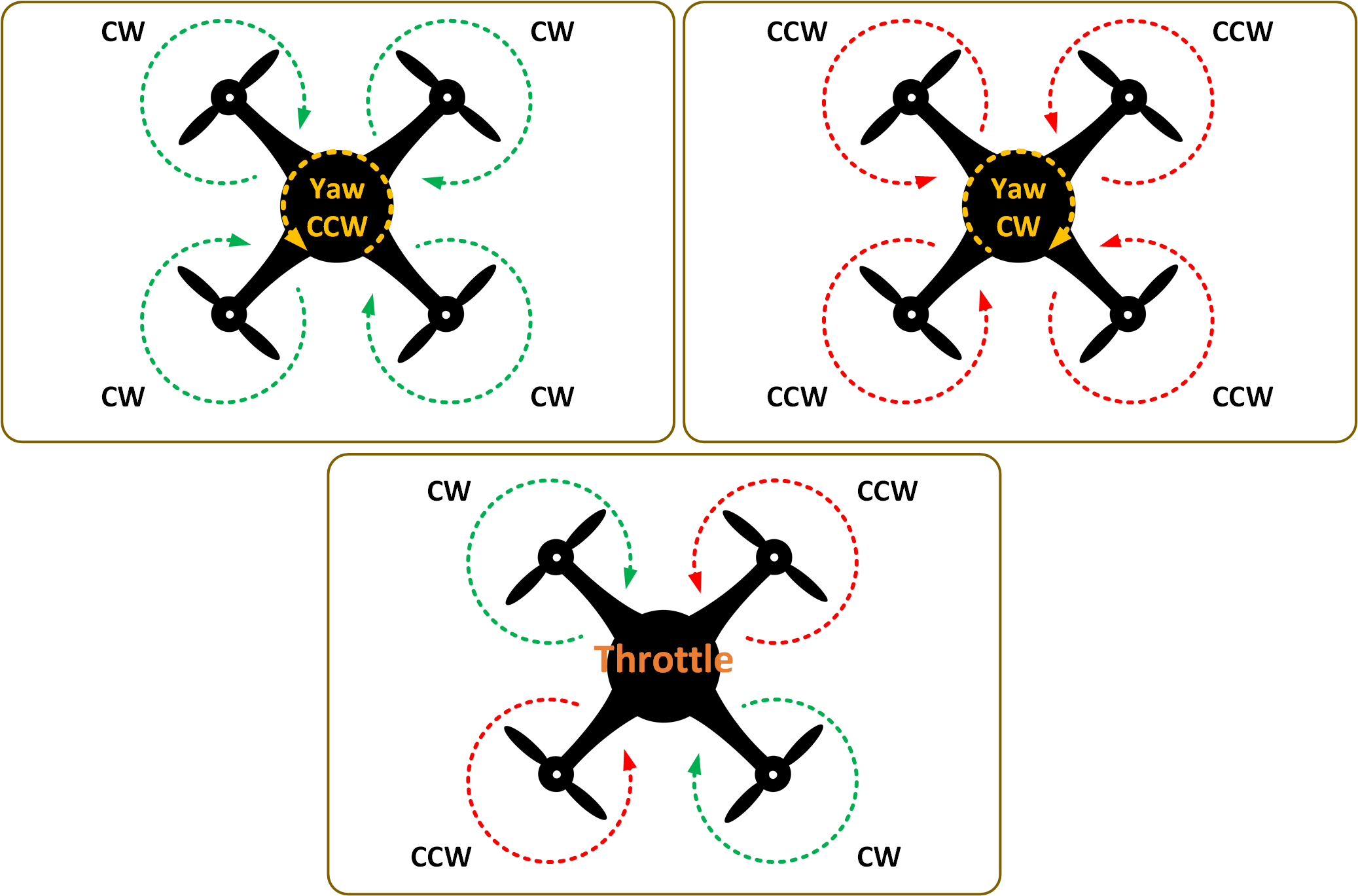}
    \caption{Yaw vs Throttle Mechanics}
    \label{fig:yaw-throttle-mechanics}
\end{figure}

\begin{figure*}[!t]
    \centering
    \includegraphics[width=1\linewidth]{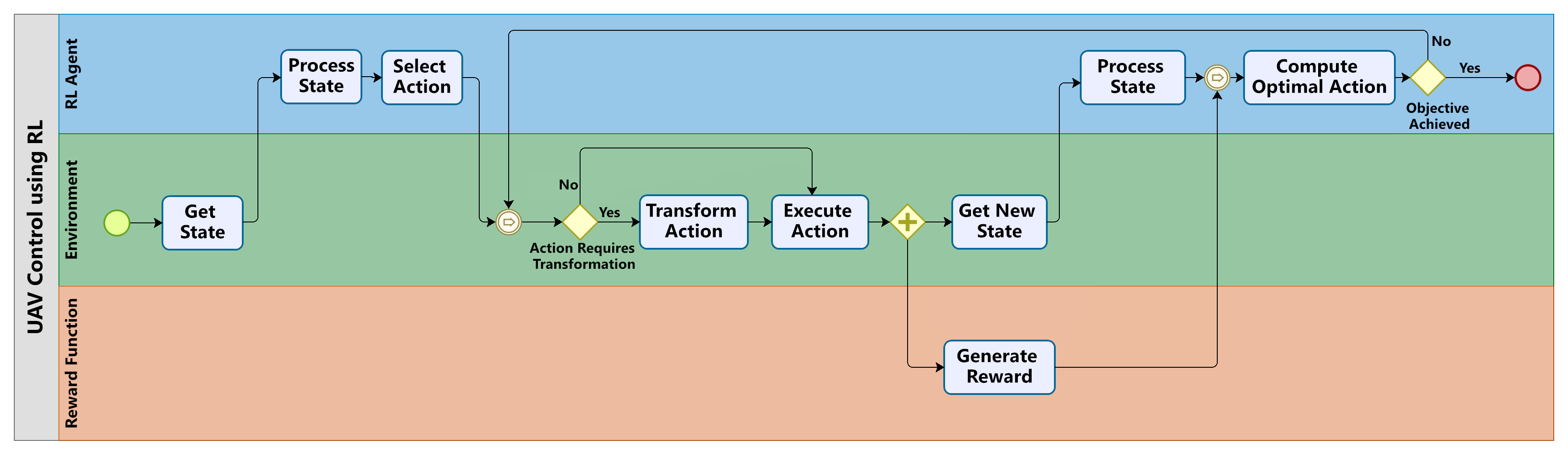}
    \caption{UAV Control using RL}
    \label{fig:uav-control}
\end{figure*}

\begin{itemize}[noitemsep,nolistsep]
    \item RGB Images: are renowned colored images where each pixel is represented in three values (Red, Green, Blue) ranging between $(0,255)$. 
    \item Depth-Map Images: contains information related to the distance of the objects from the Field Of View (FOV).  
    \item Event-Based Images: are special images that output the changes in brightness intensity instead of standard images. Event-based images are produced by an event camera, also known as Dynamic Vision Sensor (DVS).   
\end{itemize}

RGB images lack depth information, and, therefore, the agent cannot estimate how far or close the UAV is to the object leading to unexpected flying behavior. On the other hand, depth information is essential for building a successful reward function that penalizes moving closer to the objects. Some techniques used RGB images and depth-map simultaneously as input to the agent to provide more information about the environment. In contrast, event-based images data are represented as one-dimensional sequences of events over time, which is used to capture quickly changing information in the scene \citep{salvatore2020neuro}. 

Similar to cameras, distance sensors have different types, such as LiDAR, RADAR, and acoustic sensors: they estimate the distance of the surrounding objects to the UAV but require less storage size than 2D images since they do not use RGB channels.

The output generated by these devices reflects the different states that the UAV has over time, used as an input to the RL agent to make actions causing the UAV to move in different directions to avoid obstacles. The NN architecture of the RL agent is based on: 1) input type, 2) the number of inputs, and 3) the used algorithm. For example, processing RGB images or depth-map images using the DQN algorithm requires Convolutional Neural Network (CNN) followed by fully-connect layers since CNN is known for its power in processing images. In contrast, processing event-based images is performed using Spiking Neural Networks (SNN), which is designed to handle spatio-temporal data and identify spatio-temporal patterns \citep{salvatore2020neuro}.

\subsection{Path Planning}
Autonomous UAVs must have a well-defined objective before executing a flying mission. Typically, the goal is to fly from a start to a destination point, such as in delivery drones. But, the goal can also be more sophisticated, such as performing surveillance by hovering over a geographical area or participating in search and rescue operations to find a missing person.

Autonomous UAV navigation requires path planning to find the best UAV path to achieve the flying objective while avoiding obstacles. The optimal path does not always mean the shortest path or a straight line between two points; instead, the UAV aims to find a safe path while considering UAV's limited power and flying mission. 

Path planning can be divided into two main types:
\begin{itemize}[noitemsep,nolistsep]
    \item Global Path Planning: concerned with planning the path from the start point to destination point in attempt to select the optimal path.
    \item Local Path Planning: concerned with planning the local optimal waypoints in an attempt to avoid static and dynamic obstacles while considering the final destination.
\end{itemize}

Path planning can be solved using different techniques and algorithms; in this work, we focus on RL techniques used to solve global and local path planning, where the RL agent receives information from the environment and outputs the optimal waypoints according to the reward function. RL techniques can be classified according to the usage of the environment's local information 1) map-based navigation and 2) mapless navigation.

\subsubsection{Map-Based Navigation}
A UAV that adopts map-based navigation uses a representation of the environment either in 3D or 2D format. The representation might include one or more of the following about the environment: 1) the different terrains, 2) fixed-obstacles locations, and 3) charging/ground stations.

Some maps oversimplify the environment representation: the map is divided into a grid with equally-sized smaller cells that store information about the environment \citep{elnaggar2018irl, andrew2018deep, cui2021uav}. Others oversimplify the environment's structure by simplifying objects representation or by using 1D/2D to represent the environment \citep{grando2020deep, wang2020deep, liu2019distributed, yan2020flocking, williams2017information, omi2021introduction, sacharny2019optimal, yijing2017q, pham2018reinforcement, guerra2020reinforcement, bakale2020indoor} . The UAV has to plan a safe and optimal path over the cells to avoid cells containing obstacles until it reaches its destination and has to plan its stopover at the charging stations based on the battery level and path length.

In a more realistic scenario, the UAV calculates a route using the map information and the GPS signal to track the UAV's current location, starting point, and destination point. The RL agent evaluates the change in the distance and the angle between the UAV's current GPS location and target GPS location, and penalizes the reward if the difference increases or if the path is unsafe depending on the reward function (objective).

\subsubsection{Mapless Navigation}
Mapless navigation does not rely on maps; instead, it applies computer vision techniques to extract features from the environment and learn the different patterns to reach the destination, which requires computation resources that might be overwhelming for some UAVs.

Localization information of the UAV obtained by different means such as Global Positioning System (GPS) or Inertial Measurement Unit (IMU) is used in mapless navigation to plan the optimal path. DRL agent receives the scene image, the destination target, and the localization information as input and outputs the change in the UAV movements. 

For example, Zhou \textit{et al.} \citep{zhou2019vision} calculated and tracked the angle between the UAV and destination point, then encoded it with the depth image extracted from the scene and used both as a state representation for the DRL agent. Although localization information seems essential to plan the path, some techniques achieved navigation with high speed using monocular visual reactive navigation system without a GPS \citep{escobar2018r}.

\subsection{Flocking}
Although UAVs are known for performing individual tasks, they can flock to perform tasks efficiently and quickly, which requires maintaining flight formation. UAV flocking has many applications, such as search and rescue operations to cover a wide geographical area.

UAV flocking is considered a more sophisticated task than a single UAV flying mission because UAVs need to orchestrate their flight to maintain flight formation while performing other tasks such as UAV control and obstacle avoidance. Flocking can be executed using different topologies:
\begin{itemize}[noitemsep,nolistsep]
    \item Flock Centering: maintaining flight formation as suggested by Reynolds \citep{reynolds1987flocks} involves three concepts: 1) flock centering, 2) avoiding obstacles, and 3) velocity matching. This topology was applied in several research papers \citep{olfati2006flocking, lee2020autonomous, la2010flocking, jia2017three, su2009flocking}.
    \item Leader-Follower Flocking: the flock leader has its mission of reaching destination, while the followers (other UAVs) flock with the leader with a mission of maintaining distance and relative position to the leader \citep{quintero2013flocking, hung2016q}.
    \item Neighbors Flocking: close neighbors coordinate with each other, where each UAV communicates with two or more nearest neighbors to maintain flight formation by maintaining relative distance and angle to the neighbors \citep{wang2018deep, morihiro2007reinforcement, xu2018multi}.
\end{itemize}

Maintaining flight formation using RL requires communication between UAVs to learn the best policy to maintain the formation while avoiding obstacles. These RL systems can be trained using a single agent or multi-agents in centralized or distributed settings.

\subsubsection{Centralized Training}
A centralized RL agent trains a shared flocking policy to maintain the flock formation using the experience collected from all UAVs, while each UAV acts individually according to its local environment information such as obstacles. The reward function of the centralized agent can be customized to serve the flocking topology, such as flock centering or leader-follower flocking. 

Yan \textit{et al.} \citep{yan2020flocking} used Proximal Policy Optimization (PPO) algorithm to train a centralized shared flocking control policy, where each UAV flocks as close as possible to the center and decentralized execution for obstacle avoidance according to each UAV local environment information. Similarly, Hung and Givigi \citep{hung2016q} trained a leader UAV to reach a destination while avoiding obstacles and trained a shared policy for followers to flock with the leader considering the relative dynamics between the leader and the followers.

Zhao \textit{et al.} \citep{zhao2021reinforcement} used a Multi-Agent Reinforcement Learning (MARL) to train a centralized flock control policy shared by all UAVs with decentralized execution. MARL received position, speed, and flight path angle from all UAVs at each time step and tried to find the optimal flocking control policy. 

The centralized training would not produce a good generalization in neighbors flocking topology since the learned policy for one neighbor is different from other neighbors' policies due to the differences in neighbors' dynamics.

\subsubsection{Distributed Training}
UAV flocking can be trained using a distributed (decentralized) approach, where each UAV has its designated RL agent responsible for finding the optimal flock policy for the UAV. The reward function is defined to maintain distance and flying direction with other UAVs and can be customized to include further information depending on the objective. 

Flight information such as location and heading angle should be communicated to other UAVs since the RL agents are distributed, and the state representation must include information of other UAVs. Any UAV that fails to receive the information from other UAVs will cause the UAV to be isolated from the flock.

Liu \textit{et al.} \citep{liu2019distributed} proposed a decentralized DRL framework to control each UAV in a distributed setting to maximize average coverage score, geographical fairness, and minimize UAVs' energy consumption.

\section{UAV Navigation Frameworks and Simulation Software} \label{sec:frameworks-sumi-soft}
Subsection  \ref{sec:nav-frameworks} discusses and classifies the UAV navigation frameworks based on the UAV navigation objectives/sub-objectives explained in Section \ref{sec:uav-navigation}, and identifies categories such as Path Planning Frameworks, Flocking Frameworks, Energy-Aware UAV Navigation Frameworks, and others. On the other hand, Subsection \ref{sec:sim-soft} explains the simulation software’s components and the most common simulation software utilized to perform the experiments.

\subsection{UAV Navigation Frameworks} \label{sec:nav-frameworks}
In general, a software framework is a conceptual structure analogous to a blueprint used to guide the comprehending construction of the software by defining different functions and their interrelationships. By definition, RL can be considered a framework by itself. Therefore, we considered only UAV navigation frameworks that add to traditional navigation using sensors or camera data for navigation. As a result, Table \ref{tab:RL-framworks} classifies UAV frameworks based on the framework objective. The subsequent sections discuss the frameworks in more detail.

\begin{table}[!b]
    \centering
    \caption{UAV Navigation Frameworks}
    \label{tab:RL-framworks}
    \begin{tabular}{l|l}
        \hline 
        \rowcolor[HTML]{EFEFEF} \textbf{Framework Objective} & \textbf{Papers}\tabularnewline
        \hline 
        \hline 
        Energy-aware UAV Navigation & \citep{bouhamed2020ddpg, imanberdiyev2016autonomous}\tabularnewline
        \hline 
        Path Planning & \citep{walker2019deep, bouhamed2020autonomous, li2019autonomous, zhang2020iadrl, camci2019planning, yijing2017q, eslamiat2019autonomous}\tabularnewline
        \hline 
        Flocking & \citep{bouhamed2020generic, majd2018integrating}\tabularnewline
        \hline 
        Vision-Based Frameworks & \citep{andrew2018deep, he2020integrated, singla2019memory, akhloufi2019drones}\tabularnewline
        \hline 
        Transfer Learning & \citep{yoon2019hierarchical}\tabularnewline
        \hline 
    \end{tabular}
\end{table}

\subsubsection{Energy-Aware UAV Navigation Frameworks}
UAVs has limited flight time, hence operate mainly using batteries. Therefore, planning flight route and recharge stopover are crucial to reach destinations. Energy-aware UAV navigation frameworks aim to provide obstacles avoidance navigation while considering the UAV battery capacity.

Bouhamed \textit{et al.} \citep{bouhamed2020ddpg} developed a framework based on  Deep Deterministic Policy Gradient (DDPG) algorithm to guide the UAV to a target position while communicating with ground stations, allowing the UAV to recharge its battery if it drops below a specific threshold. Similarly, Imanberdiyev \textit{et al.} \citep{imanberdiyev2016autonomous} monitor battery level, rotors' condition, and sensor readings to plan the route and apply necessary route changes for required battery charging.

\subsubsection{Path Planning Frameworks}
Path planning is the process of determining the most efficient route that meets the flight objective, such as finding the shortest, fastest, or safest route. Different frameworks \citep{walker2019deep, yijing2017q} implemented a modular path planning scheme, where each module has a specialized function to achieve while exchanging data with other modules to train action selection policies and discover the optimal path.

Similarly, Li \textit{et al.} \citep{li2019autonomous} developed a four-layer framework in which each layer generates a set of objective and constraint functions. The functions are intended to serve the lower layer and consider the upper layer's objectives and constraints, with their primary goal generating trajectories.

Other frameworks suggested stage-based learning to choose actions from the desired stage depending on current environment encounters. For example, Camci and Kayacan \citep{camci2019planning} proposed learning a set of motion primitives offline, then using them online to design quick maneuvers to enable switching seamlessly between two modes: \textit{near-hover motions}, which is responsible for generating motion plans allowing a stable completion of maneuvers and \textit{swift maneuvers} to deal smoothly with abrupt inputs.

In a collaborative setting, Zhang \textit{et al.} \citep{zhang2020iadrl} suggested a coalition between Unmanned Ground Vehicle (UGV) and UAV complementing each other to reach the destination, where UAV cannot get to far locations alone due to limited battery power, and UGV cannot reach high altitude due to limited abilities.

\subsubsection{Flocking Frameworks}
UAV flocking frameworks have functionality beyond UAV navigation while maintaining flight formation. For example, Bouhamed \textit{et al.} \citep{bouhamed2020generic} presented a RL-based spatiotemporal scheduling system for autonomous UAVs. The system enables UAVs to autonomously arrange their schedules to cover the most significant number of pre-scheduled events physically and temporally spread throughout a specified geographical region and time horizon. On the other hand, Majd \textit{et al.} \citep{majd2018integrating} predicted the movement of drones and dynamic obstacles in the flying zone to generate efficient and safe routes.
\subsubsection{Vision-Based Frameworks}
Vision-Based Framework depends on UAV camera for navigation, where the images produced by the camera are used to draw on additional functionality for improved navigation. It is possible to employ frameworks that augment the agent's CNN architecture to fuse data from several sensors, use Long-Short Term Memory cells (LSTM) to maintain navigation choices, use RNN to capture the UAV states over different time steps, or pre-process images to provide more information about the environment \citep{he2020integrated, singla2019memory, he2020integrated}.
\subsubsection{Transfer Learning Frameworks}

UAVs are trained on target environments before executing the flight mission; the training is carried either in a virtual or real-world environment. The UAV requires retraining when introduced to new environments or moving from virtual training as the environments have different terrains and obstacle structures or textures. Besides, UAV training requires a long time and it is hardware resource intensive while actual UAVs have limited hardware resources. Therefore, when UAV is introduced to new environments, transfer learning frameworks reduce the training time by reusing the NN weights trained from the previous environment and retraining only parts of the agent's NN.

Yoon \textit{et al.} \citep{yoon2019hierarchical} proposed algorithm-hardware co-design, where the UAV is trained in a virtual environment, and after the UAV is deployed to a real-world environment; the agent loads the weights stored in embedded Non-Volatile Memory (eNVM), and then evaluates new actions and only trains the last few layers of CNN whose weights are stored in the on-die SRAM (Static Random Access Memory).

\subsection{Simulation Software} \label{sec:sim-soft}
The research community used different evaluation methods for autonomous UAV navigation using RL. Simulation software is used widely over actual UAVs to execute the evaluation due to the cost of the hardware (drone) in addition to the cost of replacement parts required due to UAV crashes. Comparison between simulation software is not the intended purpose, rather than making the research community aware of the most commonly used tools for evaluation as illustrated in Figure \ref{fig:simulation-software}.
3D UAV navigation simulation requires mainly three components as illustrated in Figure \ref{fig:uav-sim-soft}:

\begin{figure}[!b]
    \centering
    \includegraphics[width=1\linewidth]{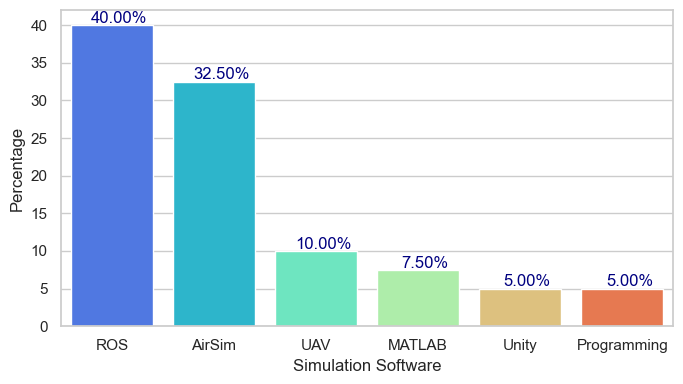}
    \caption{UAV Simulation Software Usage}
    \label{fig:simulation-software}
\end{figure}
\begin{figure}[!b]
    \centering
    \includegraphics[width=.8\linewidth]{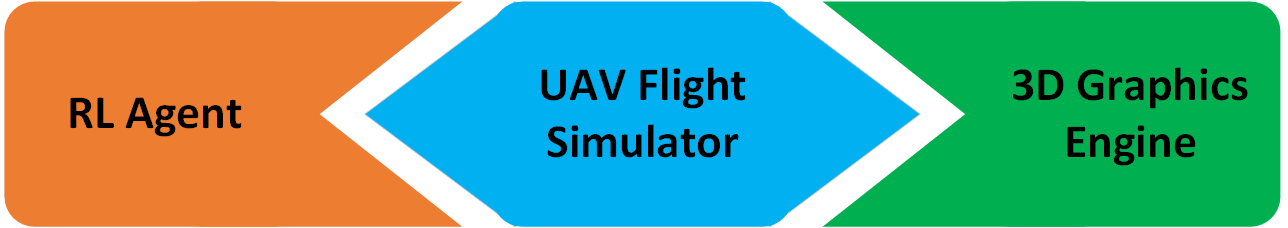}
    \caption{UAV Simulation Software Components}
    \label{fig:uav-sim-soft}
\end{figure}
\begin{itemize}[noitemsep,nolistsep]
    \item \textbf{RL Agent:} represents the RL algorithm used with all computations required to generate the reward, process the states, and compute the optimal action. RL Agent interacts directly with the \textit{UAV Flight simulator} to send/receive UAV actions/states.
    \item \textbf{UAV Flight Simulator:} responsible for simulating the UAV movements and interactions with the 3D environment, such as obtaining images from the UAV camera or reporting UAV crashes with different obstacles. Examples of UAV flight simulators are Robot Operating systems (ROS) \citep{ros2021online} and Microsoft AirSim \citep{airsim2021online}.
    \item \textbf{3D Graphics Engine:} provides a 3D graphics environment with the physics engine, which is responsible for simulating the gravity and dynamics similar to the real world. Examples of 3D graphics engines are Gazebo \citep{gazebo2021online} and Unreal Engine \citep{unrealengine2021online}.
\end{itemize}

Due to compatibility/support issues, ROS is used in conjunction with Gazebo, where AirSim uses Unreal Engine to run the simulations. However, the three components might not always be present, especially if the simulation software has internal modules or plugins that provide the required functionality, such as MATLAB.
\section{Reinforcement Learning Algorithms Classification} \label{sec:rl-algorithms}
The previous sections discussed the UAV navigation tasks and frameworks without elaborating on RL algorithms. However, to choose a suitable algorithm for the application environment and the navigation task, the comprehension of RL algorithms and their characteristics is necessary. For example, the DQN algorithm and its variations can be used for UAV navigation tasks that use simple movement actions (UP, DOWN, LEFT, RIGHT, FORWARD) since they are discrete. Therefore, this section examines RL algorithms and their characteristics.

AlMahamid and Grolinger \citep{almahamid2021reinforcement} categorized RL algorithms into three main categories according to the number of states and the type of actions: 1) limited number of states and discrete actions, 2) unlimited number of states and discrete actions, and 3) unlimited number of states and continuous actions. We extend this with sub-classes, analyze more than 50 RL algorithms, and examine their use in UAV navigation. Table \ref{tab:rl-algorithms} classifies all RL algorithms found in UAV Navigation studies and includes other prominent RL algorithms to show the intersection between RL and UAV navigation. Furthermore, this section discusses algorithms characteristics and highlights RL applications in different UAV navigation studies. Note that Table \ref{tab:rl-algorithms} includes many algorithms, but only the most prominent ones are discussed in the following subsections. 

\begin{table*}[!t]
\centering
\caption{RL Algorithms usage and classification}
\label{tab:rl-algorithms}

\begin{tabular}{>{\centering}m{0.04\linewidth}|>{\centering}m{0.04\linewidth}|>{\centering}m{0.03\linewidth}|>{\centering}m{0.15\linewidth}|>{\centering}m{0.05\linewidth}|>{\centering}m{0.05\linewidth}|>{\centering}m{0.05\linewidth}|>{\centering}m{0.05\linewidth}|>{\centering}m{0.05\linewidth}|>{\raggedright}m{0.28\linewidth}}
\hline 
\rowcolor[HTML]{EFEFEF} \textbf{State} & \textbf{Action} & \textbf{Class} & \textbf{Algorithm} & \textbf{On/Off\linebreak Policy} & \textbf{Actor-Critic} & \textbf{Multi-Thread} & \hspace{0pt}\textbf{Distributed} & \textbf{Multi-Agent} & \textbf{Usage}\tabularnewline
\hline 
\hline 
\multirow{2}{0.04\linewidth}{\crotate{0.9}{Limited}} & \multirow{2}{0.04\linewidth}{\crotate{0.95}{Discrete}} & \multirow{2}{0.03\linewidth}{\crotate{0.85}{Simple RL}} & Q-Learning \cite{watkins1992q} & - & No & No & No & No & \cite{bouhamed2020generic, yu2019navigation, li2018path, sacharny2019optimal, karthik2020reinforcement, pham2018reinforcement, guerra2020reinforcement, bouhamed2020uav, kulkarni2020uav, cui2021uav, fotouhi2021deep, greatwood2019reinforcement}\tabularnewline
\cline{4-10} \cline{5-10} \cline{6-10} \cline{7-10} \cline{8-10} \cline{9-10} \cline{10-10} 
 &  &  & SARSA \cite{rummery1994line}  & - & No & No & No & No & -\tabularnewline
\hline 
\multirow{49}{0.04\linewidth}{\crotate{1}{Unlimited}} & \multirow{22}{0.04\linewidth}{\crotate{1}{Discrete}} & \multirow{6}{0.03\linewidth}{\crotate{1.5}{DQN Variations}} & DQN \cite{mnih2013playing} & Off & No & No & No & No & \cite{salvatore2020neuro, huang2019autonomous, shin2019automatic, chen2020collision, abedin2020data, camci2020deep, huang2019deep, cetin2019drone, williams2017information, wu2018navigating, zhou2018neural, camci2019planning, yijing2017q, walvekar2019vision, viseras2021wildfire, eslamiat2019autonomous, fotouhi2021deep, akhloufi2019drones, bakale2020indoor, camci2019learning, madridano2021software, bonnet2019uav}\tabularnewline
\cline{4-10} \cline{5-10} \cline{6-10} \cline{7-10} \cline{8-10} \cline{9-10} \cline{10-10} 
 &  &  & Double DQN \cite{VanHasselt2016} & Off & No & No & No & No & \cite{zhou2020efficient, shin2019automatic, anwar2020autonomous, yang2020autonomous, cetin2019drone, yoon2019hierarchical, anwar2018navren, polvara2018toward, polvara2019autonomous, fotouhi2021deep, munoz2019deep}\tabularnewline
\cline{4-10} \cline{5-10} \cline{6-10} \cline{7-10} \cline{8-10} \cline{9-10} \cline{10-10} 
 &  &  & Dueling DQN \cite{Wang2016} & Off & No & No & No & No & \cite{shin2019automatic, cetin2019drone}\tabularnewline
\cline{4-10} \cline{5-10} \cline{6-10} \cline{7-10} \cline{8-10} \cline{9-10} \cline{10-10} 
 &  &  & DRQN \cite{hausknecht2015deep} & Off & No & No & No & No & \cite{andrew2018deep, singla2019memory, peake2020wilderness, tong2021uav}\tabularnewline
\cline{4-10} \cline{5-10} \cline{6-10} \cline{7-10} \cline{8-10} \cline{9-10} \cline{10-10} 
 &  &  & DD-DQN \cite{Wang2016} & Off & No & No & No & No & \cite{shin2019automatic, villanueva2019deep}\tabularnewline
\cline{4-10} \cline{5-10} \cline{6-10} \cline{7-10} \cline{8-10} \cline{9-10} \cline{10-10} 
 &  &  & DD-DRQN & Off & No & No & No & No & -\tabularnewline
\cline{3-10} \cline{4-10} \cline{5-10} \cline{6-10} \cline{7-10} \cline{8-10} \cline{9-10} \cline{10-10} 
 &  & \multirow{6}{0.03\linewidth}{\crotate{1.5}{Distributional DQN}} & Noisy DQN \cite{fortunato2017noisy} & Off & No & No & No & No & -\tabularnewline
\cline{4-10} \cline{5-10} \cline{6-10} \cline{7-10} \cline{8-10} \cline{9-10} \cline{10-10} 
 &  &  & C51-DQN \cite{bellemare2017distributional}  & Off & No & No & No & No & -\tabularnewline
\cline{4-10} \cline{5-10} \cline{6-10} \cline{7-10} \cline{8-10} \cline{9-10} \cline{10-10} 
 &  &  & QR-DQN \cite{dabney2018distributional} & Off & No & No & No & No & -\tabularnewline
\cline{4-10} \cline{5-10} \cline{6-10} \cline{7-10} \cline{8-10} \cline{9-10} \cline{10-10} 
 &  &  & IQN \cite{dabney2018implicit} & Off & No & No & No & No & -\tabularnewline
\cline{4-10} \cline{5-10} \cline{6-10} \cline{7-10} \cline{8-10} \cline{9-10} \cline{10-10} 
 &  &  & Rainbow DQN \cite{hessel2018rainbow} & Off & No & No & No & No & -\tabularnewline
\cline{4-10} \cline{5-10} \cline{6-10} \cline{7-10} \cline{8-10} \cline{9-10} \cline{10-10} 
 &  &  & FQF \cite{yang2019fully} & Off & No & No & No & No & -\tabularnewline
\cline{3-10} \cline{4-10} \cline{5-10} \cline{6-10} \cline{7-10} \cline{8-10} \cline{9-10} \cline{10-10} 
 &  & \multirow{4}{0.03\linewidth}{\crotate{1.5}{Distributed DQN}} & R2D2 \cite{kapturowski2018recurrent} & Off & No & No & Yes & No & -\tabularnewline
\cline{4-10} \cline{5-10} \cline{6-10} \cline{7-10} \cline{8-10} \cline{9-10} \cline{10-10} 
 &  &  & Ape-X DQN \cite{horgan2018distributed} & Off & No & No & Yes & No & -\tabularnewline
\cline{4-10} \cline{5-10} \cline{6-10} \cline{7-10} \cline{8-10} \cline{9-10} \cline{10-10} 
 &  &  & NGU \cite{badia2020never} & Off & No & No & Yes & No & -\tabularnewline
\cline{4-10} \cline{5-10} \cline{6-10} \cline{7-10} \cline{8-10} \cline{9-10} \cline{10-10} 
 &  &  & Agent57 \cite{badia2020agent57} & Off & No & No & Yes & No & -\tabularnewline
\cline{3-10} \cline{4-10} \cline{5-10} \cline{6-10} \cline{7-10} \cline{8-10} \cline{9-10} \cline{10-10} 
 &  & \multirow{6}{0.03\linewidth}{\crotate{2}{Deep SARSA Variations}} & Deep SARSA \cite{zhao2016deep} & On & No & No & No & No & -\tabularnewline
\cline{4-10} \cline{5-10} \cline{6-10} \cline{7-10} \cline{8-10} \cline{9-10} \cline{10-10} 
 &  &  & Double SARSA & On & No & No & No & No & -\tabularnewline
\cline{4-10} \cline{5-10} \cline{6-10} \cline{7-10} \cline{8-10} \cline{9-10} \cline{10-10} 
 &  &  & Dueling SARSA & On & No & No & No & No & -\tabularnewline
\cline{4-10} \cline{5-10} \cline{6-10} \cline{7-10} \cline{8-10} \cline{9-10} \cline{10-10} 
 &  &  & DR-SARSA & On & No & No & No & No & -\tabularnewline
\cline{4-10} \cline{5-10} \cline{6-10} \cline{7-10} \cline{8-10} \cline{9-10} \cline{10-10} 
 &  &  & DD-SARSA & On & No & No & No & No & -\tabularnewline
\cline{4-10} \cline{5-10} \cline{6-10} \cline{7-10} \cline{8-10} \cline{9-10} \cline{10-10} 
 &  &  & DD-DR-SARSA & On & No & No & No & No & -\tabularnewline
\cline{2-10} \cline{3-10} \cline{4-10} \cline{5-10} \cline{6-10} \cline{7-10} \cline{8-10} \cline{9-10} \cline{10-10} 
 & \multirow{27}{0.04\linewidth}{\crotate{1}{Continous}} & \multirow{5}{0.03\linewidth}{\crotate{1}{Policy Based}} & REINFORCE \cite{williams1992simple} & On & No & No & No & No & -\tabularnewline
\cline{4-10} \cline{5-10} \cline{6-10} \cline{7-10} \cline{8-10} \cline{9-10} \cline{10-10} 
 &  &  & TPRO \cite{schulman2015trust} & On & No & No & No & No & \cite{koch2019reinforcement}\tabularnewline
\cline{4-10} \cline{5-10} \cline{6-10} \cline{7-10} \cline{8-10} \cline{9-10} \cline{10-10} 
 &  &  & PPO \cite{schulman2017proximal} & On & No & No & No & No & \cite{morad2021embodied, yan2020flocking, zhang2020iadrl, hasanzade2021dynamically, wang2021pretrained, hodge2021deep, deshpande2020developmental, maxey2019navigation, koch2019reinforcement}\tabularnewline
\cline{4-10} \cline{5-10} \cline{6-10} \cline{7-10} \cline{8-10} \cline{9-10} \cline{10-10} 
 &  &  & PPG \cite{cobbe2020phasic} & Off & No & No & No & No & -\tabularnewline
\cline{4-10} \cline{5-10} \cline{6-10} \cline{7-10} \cline{8-10} \cline{9-10} \cline{10-10} 
 &  &  & SVPG \cite{liu2017stein} & Off & No & No & No & No & -\tabularnewline
\cline{3-10} \cline{4-10} \cline{5-10} \cline{6-10} \cline{7-10} \cline{8-10} \cline{9-10} \cline{10-10} 
 &  & \multirow{8}{0.03\linewidth}{\crotate{1.6}{Actor-Critic}} & SLAC \cite{lee2019stochastic} & Off & Yes & No & No & No & -\tabularnewline
\cline{4-10} \cline{5-10} \cline{6-10} \cline{7-10} \cline{8-10} \cline{9-10} \cline{10-10} 
 &  &  & ACE \cite{zhang2019ace} & Off & Yes & Yes & No & No & -\tabularnewline
\cline{4-10} \cline{5-10} \cline{6-10} \cline{7-10} \cline{8-10} \cline{9-10} \cline{10-10} 
 &  &  & DAC \cite{zhang2019dac} & Off & Yes & No & No & No & -\tabularnewline
\cline{4-10} \cline{5-10} \cline{6-10} \cline{7-10} \cline{8-10} \cline{9-10} \cline{10-10} 
 &  &  & DPG \cite{silver2014deterministic} & Off & Yes & No & No & No & \cite{li2019optimal}\tabularnewline
\cline{4-10} \cline{5-10} \cline{6-10} \cline{7-10} \cline{8-10} \cline{9-10} \cline{10-10} 
 &  &  & RDPG \cite{heess2015memory} & Off & Yes & No & No & No & -\tabularnewline
\cline{4-10} \cline{5-10} \cline{6-10} \cline{7-10} \cline{8-10} \cline{9-10} \cline{10-10} 
 &  &  & DDPG \cite{lillicrap2015continuous} & Off & Yes & No & No & No & \cite{bouhamed2020ddpg, wang2018deep, wang2020two, bouhamed2020uav, bouhamed2020autonomous, li2019autonomous, grando2020deep, wang2020deep, liu2019distributed, he2020integrated, zhou2019vision, doukhi2021deep, koch2019reinforcement, lee2018vision}\tabularnewline
\cline{4-10} \cline{5-10} \cline{6-10} \cline{7-10} \cline{8-10} \cline{9-10} \cline{10-10} 
 &  &  & TD3 \cite{fujimoto2018addressing} & Off & Yes & No & No & No & \cite{omi2021introduction}\tabularnewline
\cline{4-10} \cline{5-10} \cline{6-10} \cline{7-10} \cline{8-10} \cline{9-10} \cline{10-10} 
 &  &  & SAC \cite{haarnoja2018soft} & Off & Yes & No & No & No & \cite{grando2020deep}\tabularnewline
\cline{3-10} \cline{4-10} \cline{5-10} \cline{6-10} \cline{7-10} \cline{8-10} \cline{9-10} \cline{10-10} 
 &  & \multirow{14}{0.03\linewidth}{\crotate{5}{Multi-Agent and Distributed Actor-Critic}} & Ape-X DPG \cite{horgan2018distributed} & Off & Yes & No & Yes & No & -\tabularnewline
\cline{4-10} \cline{5-10} \cline{6-10} \cline{7-10} \cline{8-10} \cline{9-10} \cline{10-10} 
 &  &  & D4PG \cite{barth2018distributed} & Off & Yes & No & Yes & Yes & -\tabularnewline
\cline{4-10} \cline{5-10} \cline{6-10} \cline{7-10} \cline{8-10} \cline{9-10} \cline{10-10} 
 &  &  & A2C \cite{mnih2016asynchronous} & On & Yes & Yes & Yes & No & \cite{lee2020autonomous, peake2020wilderness}\tabularnewline
\cline{4-10} \cline{5-10} \cline{6-10} \cline{7-10} \cline{8-10} \cline{9-10} \cline{10-10} 
 &  &  & DPPO \cite{heess1707emergence} & On & No & No & Yes & Yes & -\tabularnewline
\cline{4-10} \cline{5-10} \cline{6-10} \cline{7-10} \cline{8-10} \cline{9-10} \cline{10-10} 
 &  &  & A3C \cite{mnih2016asynchronous} & On & Yes & Yes & No & Yes & \cite{wang2020deep}\tabularnewline
\cline{4-10} \cline{5-10} \cline{6-10} \cline{7-10} \cline{8-10} \cline{9-10} \cline{10-10} 
 &  &  & PAAC \cite{alfredo2017efficient} & On & Yes & Yes & No & No & -\tabularnewline
\cline{4-10} \cline{5-10} \cline{6-10} \cline{7-10} \cline{8-10} \cline{9-10} \cline{10-10} 
 &  &  & ACER \cite{wang2016sample} & Off & Yes & Yes & No & No & -\tabularnewline
\cline{4-10} \cline{5-10} \cline{6-10} \cline{7-10} \cline{8-10} \cline{9-10} \cline{10-10} 
 &  &  & Reactor \cite{gruslys2017reactor} & Off & Yes & Yes & No & No & -\tabularnewline
\cline{4-10} \cline{5-10} \cline{6-10} \cline{7-10} \cline{8-10} \cline{9-10} \cline{10-10} 
 &  &  & ACKTR \cite{wu2017scalable} & On & Yes & Yes & No & No & -\tabularnewline
\cline{4-10} \cline{5-10} \cline{6-10} \cline{7-10} \cline{8-10} \cline{9-10} \cline{10-10} 
 &  &  & MADDPG \cite{lowe2017multi} & Off & Yes & No & No & Yes & -\tabularnewline
\cline{4-10} \cline{5-10} \cline{6-10} \cline{7-10} \cline{8-10} \cline{9-10} \cline{10-10} 
 &  &  & MATD3 \cite{ackermann2019reducing} & Off & Yes & No & No & Yes & -\tabularnewline
\cline{4-10} \cline{5-10} \cline{6-10} \cline{7-10} \cline{8-10} \cline{9-10} \cline{10-10} 
 &  &  & MAAC \cite{iqbal2019actor} & Off & Yes & No & No & Yes & \cite{fan2020prioritized}\tabularnewline
\cline{4-10} \cline{5-10} \cline{6-10} \cline{7-10} \cline{8-10} \cline{9-10} \cline{10-10} 
 &  &  & IMPALA \cite{espeholt2018impala} & Off & Yes & Yes & Yes & Yes & -\tabularnewline
\cline{4-10} \cline{5-10} \cline{6-10} \cline{7-10} \cline{8-10} \cline{9-10} \cline{10-10} 
 &  &  & SEED \cite{espeholt2019seed} & Off & Yes & Yes & Yes & Yes & -\tabularnewline
\hline 
\end{tabular}

\end{table*}

\subsection{Limited States and Discrete Actions} \label{LimitedStatesDiscreteActions}
Generally, simple environments have a limited number of states and the agent transitions between states by executing discrete (limited number) actions. For example, in a tic-tac-toe game, the agent has a predefined set of two actions \textit{X} or \textit{O} that are used to update the nine boxes constituting the predefined set of known states. Q-Learning \citep{watkins1992q} and State–Action–Reward–State–Action (SARSA) \citep{rummery1994line} algorithms can be applied to environments with a limited number of states and discrete actions, where they maintain a Q-Table with all possible states and actions while iteratively updating the Q-values for each state-action pair to find the optimal policy.

SARSA is similar to Q-Learning except to update the current $\bm{Q(s,a)}$ value it computes the next state-action $\bm{Q(s^\prime,a^\prime)}$ by executing the next action $\bm{a^\prime}$ \citep{zhao2016deep}. In contrast, Q-learning updates the current $\bm{Q(s,a)}$ value by computing the next state-action $\bm{Q(s^\prime,a^\prime)}$ using the Bellman equation since the next action is unknown, and takes a greedy action by selecting the action that maximizes the reward \citep{zhao2016deep}.
\subsection{Unlimited States and Discrete Actions}
An RL agent uses Deep Neural Network (DNN) - usually a CNN, in complex environments such as the pong game, where the states are unlimited and the actions are discrete (UP, DOWN). The deep agent/DNN processes the environment's state as an input and outputs the Q-values of the available actions. The following subsections discuss the different algorithms that can be used in this type of the environment, such as DQN, Deep SARSA, and their variations \citep{almahamid2021reinforcement}.

\subsubsection{Deep Q-Networks Variations}
\begin{figure}[!t]
    \centering
    \includegraphics[width=1\linewidth]{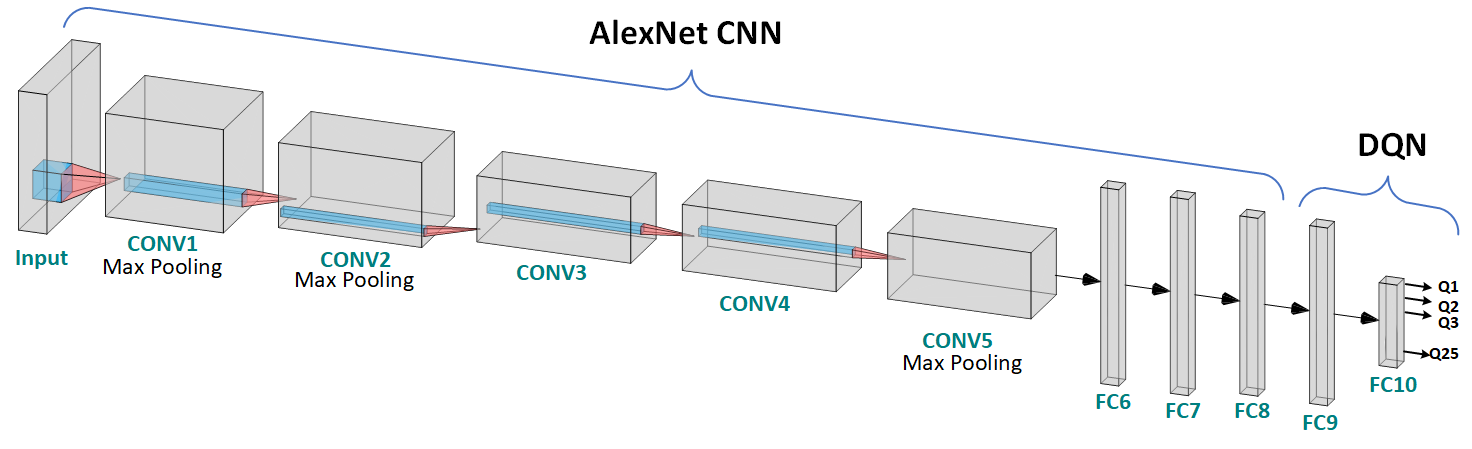}
    \caption{DQN using AlexNet CNN}
    \label{fig:dqn}
\end{figure}

Deep Q-Learning, also known as Deep Q-Network (DQN)\citep{mnih2013playing}, is a primary method used in settings with an unlimited number of states and discrete actions, and it serves as an inspiration for other algorithms used for the same goal. As illustrated in Figure \ref{fig:dqn} \citep{anwar2020autonomous}, DQN architecture frequently employs convolutional and pooling layers, followed by fully connected layers that provide Q-values corresponding to the number of actions. A significant disadvantage of the DQN algorithm is that it overestimates the action-value (Q-value), with the agent selecting the actions with the highest Q-value, which may not be the optimal action \citep{VanHasselt2010}. 

Double DQN solves the overestimation issue in DQN by using two networks. The first network, known as the Policy Network, optimizes the Q-value, while the second network, known as the Target Network, is a clone of the Policy Network and is used to generate the estimated Q-value \citep{VanHasselt2016}. After a specified number of time steps, the parameters of the target network network are updated by copying the policy network parameters instead of performing backpropagation. 

\begin{figure}[!t]
    \centering
    \includegraphics[width=1\linewidth]{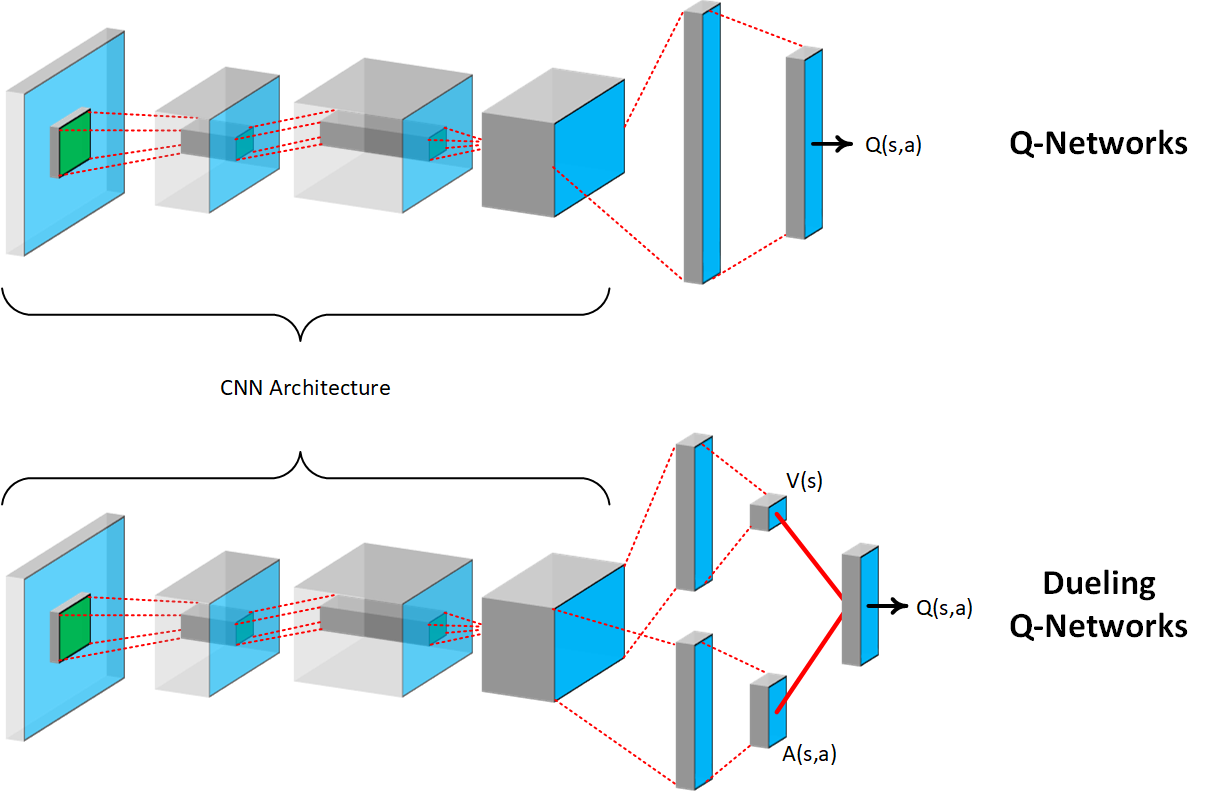}
    \caption{DQN vs. Dueling DQN}
    \label{fig:Dueling-DQN}
\end{figure}

Dueling DQN, as depicted in Figure \ref{fig:Dueling-DQN} \citep{Wang2016}, is a further enhancement to DQN. To improve Q-value evaluation, Dueling DQN employs the following functions in place the Q-value function: 

\begin{itemize}[noitemsep,nolistsep]
    \item The State-Value function $\bm{V(s)}$ quantifies how desirable it is for an agent to be in a state $\bm{s}$. 
    \item The Advantage-Value function $\bm{A(s, a)}$ assesses the superiority of the selected action in a given state $\bm{s}$ over other actions. 
\end{itemize}

The two functions depicted in Figure \ref{fig:Dueling-DQN} are integrated using a custom aggregation layer to generate an estimate of the state-action value function \citep{Wang2016}. The aggregation layer has the same value as the sum of the two values produced by the two functions: 

\begin{equation}
    \label{eq:dueling-dqn}
    Q(s,a) = V(s) + \big( A(s,a)  -\frac{1}{|\mathcal{A}|} \sum_{a^\prime} A(s,a) \big)
\end{equation}

The term $\bm{\frac{1}{|\mathcal{A}|} \sum_{a^\prime} A(s, a)}$ denotes the mean, whereas $\bm{|\mathcal{A}|}$ denotes vector $\bm{A}$ length. This assists the \textit{identifiability} problem while having no effect on the relative rank of the $A$ (and thus Q) values. This also improves the optimization because the advantage function only needs to change as fast as the mean \citep{Wang2016}.

As explained by Wang \textit{et al.} \citep{Wang2016}, Double Dueling DQN (DD-DQN) extends DQN by combining Dueling DQN and Double DQN to determine the optimal Q-value, with the output of Dueling DQN passed to Double DQN.

The Deep Recurrent Q-Network (DRQN) \citep{hausknecht2015deep} algorithm is a DQN variation, using a recurrent LSTM layer in place of the first fully connected layer. This changes the input from a single environment state to to a group of states as a single input, which aids in the integration of information over time \citep{hausknecht2015deep}. The techniques of doubling and dueling can be utilized independently or in combination with a recurrent neural network. 

\subsubsection{Distributional DQN}
The goal of distributional Q-learning is to obtain a more accurate representation of the distribution of observed rewards. Fortunato \textit{et al.} \citep{fortunato2017noisy} introduced NoisyNet, a deep reinforcement learning agent that uses gradient descent to learn parametric noise added to the network weights, and demonstrated how the agent's policy's induced stochasticity can be used to aid efficient exploration \citep{fortunato2017noisy}.

\noindent
\textbf{Categorical Deep Q-Networks (C51-DQN)} \citep{bellemare2017distributional} applied a distributional perspective using Wasserstein metric to the random return received by Bellman's equation to approximate value distributions instead of the value function. The algorithm first performs a heuristic projection step and then minimizes the Kullback-Leibler (KL) divergence between the projected Bellman update and the prediction \citep{bellemare2017distributional}.

\noindent
\textbf{Quantile Regression Deep Q-Networks (QR-DQN)} \citep{dabney2018distributional} performs a distributional reinforcement learning over the Wasserstein metric in a stochastic approximation setting. Using Wasserstein distance, the target distribution is minimized by stochastically adjusting the distributions’ locations using quantile regression \citep{dabney2018distributional}. QR-DQN assigns fixed, uniform probabilities to $N$ adjustable locations and minimizes the quantile Huber loss between the Bellman updated distribution and current return distribution \citep{yang2019fully}, whereas C51-DQN uses $N$ fixed locations ($N=51$) for distribution approximation and adjusts the locations probabilities \citep{dabney2018distributional}.

\noindent
\textbf{Implicit Quantile Networks (IQN)} \citep{dabney2018implicit} incorporates QR-DQN \citep{dabney2018distributional} to learn full quantile function controlled by the size of the network and the amount of training, in contrast to QR-DQN  quantile function that learns a discrete set of quantiles dependent on the number of quantiles output \citep{dabney2018implicit}. IQN distribution function assumes the base distribution to be non-uniform and reparameterizes samples from a base distribution to the respective quantile values of a target distribution.

\noindent
\textbf{Rainbow DQN} \citep{hessel2018rainbow} combines several improvements of the traditional DQN algorithm into a single algorithm, such as \textbf{1)} addressing the overestimation bias, \textbf{2)} using Prioritized Experience Replay (PER) \citep{schaul2015prioritized}, \textbf{3)} using Dueling DQN \citep{Wang2016}, \textbf{4)} shifting the bias-variance trade-off and propagating newly observed rewards faster to earlier visited states as implemented in A3C \citep{mnih2016asynchronous}, \textbf{5)} learning a distributional reinforcement learning instead of the expected return similar to C51-DQN \citep{bellemare2017distributional}, and \textbf{6)} implementing stochastic network layers using Noisy DQN \citep{fortunato2017noisy}.

Yang \textit{et al.} \citep{yang2019fully} proposed \textbf{Fully parameterized Quantile Function (FQF)} for distributional RL providing full parameterization for both quantile fractions and corresponding quantile values. In contrast, QR-DQN \citep{dabney2018distributional} and IQN \citep{dabney2018implicit} only parameterize the corresponding quantile values, while quantile fractions are either fixed or sampled \citep{yang2019fully}.

FQF for distributional RL uses two networks: 1) quantile value network that maps quantile fractions to corresponding quantile values, and 2) fraction proposal network that generates quantile fractions for each state-action pair with the goal of distribution approximation while minimizing the 1-Wasserstein distance between the approximated and actual distribution \citep{yang2019fully}.

\subsubsection{Distributed DQN}
\begin{figure}[!b]
    \centering
    \includegraphics[width=.9\linewidth]{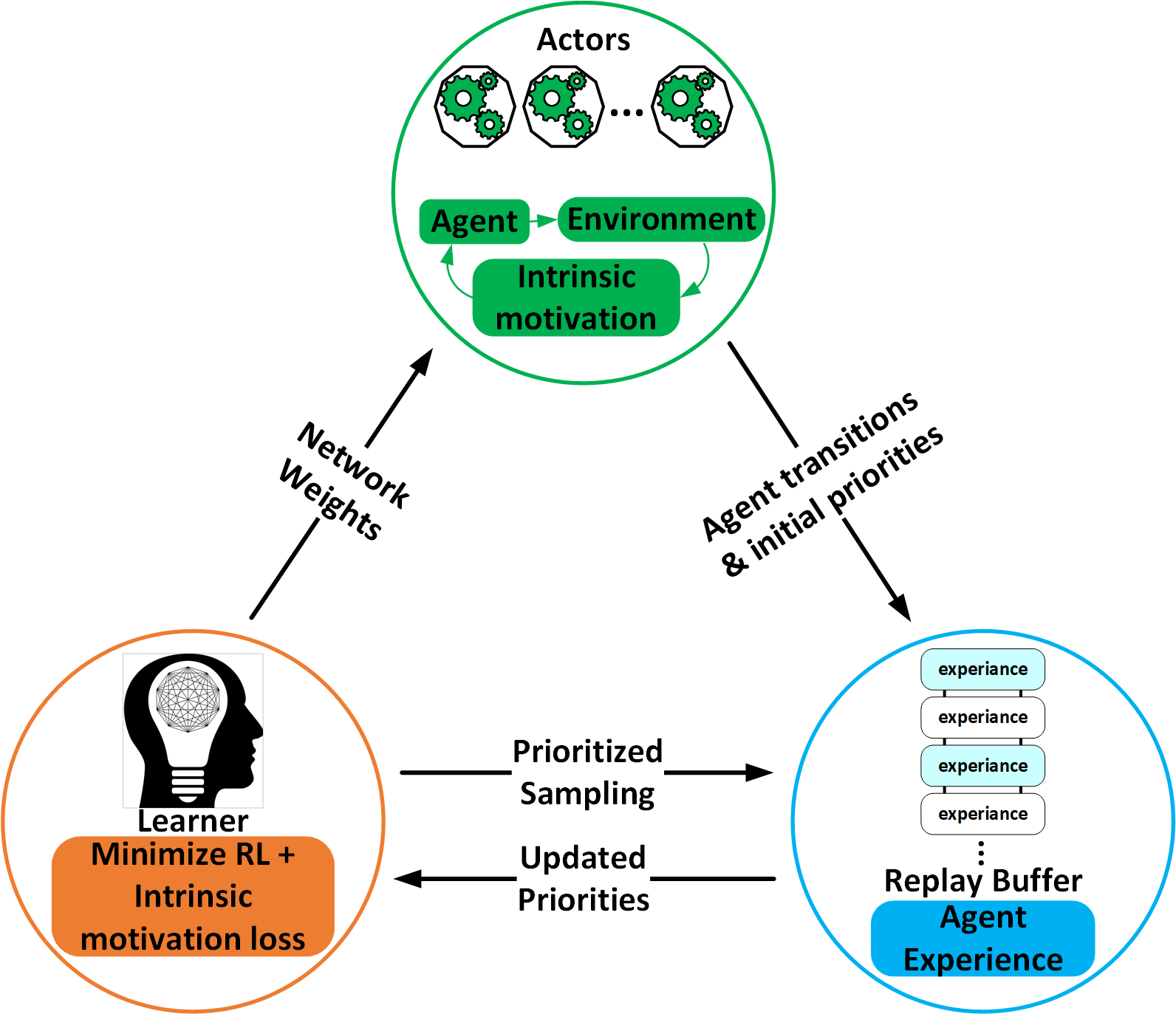}
    \caption{Distributed DRL agent scheme}
    \label{fig:distributed-drl}
\end{figure}

Distributed DRL architecture used by different RL algorithms as depicted in Figure \ref{fig:distributed-drl} \citep{badia2020agent57} aims to decouple acting from learning in distributed settings relaying on prioritized experience replay to focus on the significant experiences generated by actors. The actors share the same NN and replay experience buffer, where they interact with the environment and store their experiences in the shared replay experience buffer. On the other hand, the learner replays prioritized experiences from the shared experience buffer and updates the learner NN accordingly \citep{horgan2018distributed}. In theory, both acting and learning can be distributed across multiple workers or running on the same machine \citep{horgan2018distributed}.

\noindent
\textbf{Ape-X DQN} \citep{horgan2018distributed}, based on the Ape-X framework, was the first algorithm to suggest distributed DRL, which was later extended by Recurrent Replay Distributed DQN (R2D2) \citep{kapturowski2018recurrent} with two main differences: 1) R2D2 adds an LSTM layer after the convolutional stack to overcome partial observability, and 2) it trains a recurrent neural network from randomly sampled replay sequences using the “burn-in” strategy, which produces a start state through using a portion of the replay sequence and updates the network only on the remaining part of the sequence \citep{kapturowski2018recurrent}.

\noindent
\textbf{Never Give Up (NGA)} \citep{badia2020never} is another algorithm that combines R2D2 architecture with a novel approach that encourages the agent to learn exploratory strategies throughout the training process using a compound intrinsic reward consisting of two modules:
\begin{itemize}[noitemsep,nolistsep]
    \item Life-long novelty module uses Random Network Distillation (RND) \citep{burda2018exploration}, which consists of two networks used to generate an intrinsic reward: 1) target network, and 2) prediction network. This mechanism is known as curiosity because it motivates the agent to explore the environment by going to novel or unfamiliar states. 
    \item Episodic novelty module uses dynamically-sized episodic memory $M$ that stores the controllable states in an online fashion, then turns state-action counts into a bonus reward, where the count is computed using the $k$-nearest neighbors.
\end{itemize}

While NGA uses intrinsic reward to promote exploration, it promotes exploitation by generating extrinsic reward using the Universal Value Function Approximator (UVFA). NGA uses conditional architecture with shared weights to learn a family of policies that separate exploration and exploitation \citep{badia2020never}.

\noindent
\textbf{Agent57} \citep{badia2020agent57} is the first RL algorithm that outperforms the human benchmark on all 57 games of Atari 2600. Agent57 implements NGA algorithms with the main difference of applying an adaptive mechanism for exploration-exploitation trade-off and utilizes parameterization of the architecture that allows for more consistent and stable learning \citep{badia2020agent57}.

\subsubsection{Deep SARSA}
SARSA is based on Q-learning and is designed for situations with limited states and discrete actions, as explained in subsection \ref{LimitedStatesDiscreteActions}. Deep SARSA \citep{zhao2016deep} uses a deep neural network similar to DQN and has the same extensions: Double SARSA, Dueling SARSA, Double Dueling SARSA (DD-SARSA), Deep Recurrent SARSA (DR-SARSA), and Double Dueling Deep Recurrent (DD-DR-SARSA). The main difference compared to DQN is that Deep SARSA computes $\bm{Q(s^\prime,a^\prime)}$ by taking the next action $\bm{a^\prime}$, which is necessary to determine the current state-action $\bm{Q(s,a)}$ rather than taking a greedy action that maximizes the reward.

\subsection{Unlimited States and Continuous Actions}
While discrete actions are adequate to drive a car or unmanned aerial vehicle in a simulated environment, they do not enable realistic movements in real-world scenarios. Continuous actions specify the quantity of movement in various directions, and the agent does not select from a predetermined set of actions. For instance, a realistic UAV movement defines the amount of roll, pitch, yaw, and throttle changes necessary to navigate the environment while avoiding obstacles, as opposed to flying the UAV in preset directions: forward, left, right, up, and down \citep{almahamid2021reinforcement}. 

Continuous action space demands learning a parameterized policy $\bm{\pi_\theta}$ that maximizes the expected summation of discounted rewards since it is not feasible to determine the action-value $\bm{Q(s,a)}$ for all continuous actions in all distinct states. Learning a parameterized policy $\bm{\pi_\theta}$ is considered a maximization problem, which can be handled using gradient descent methods to get the optimal $\bm{\theta}$ in the following manner \citep{almahamid2021reinforcement}:

\begin{equation}
    \label{eq:gradient-ascent}
    \theta_{t+1} = \theta_{t} + \alpha\nabla J(\theta_{t})
\end{equation}

\noindent Here, $\bm{\nabla}$ is the gradient and $\bm{\alpha}$ is the learning rate. 

The goal of the reward function $\bm{J}$ is to maximize the expected reward applying the following parameterized policy $\pi_\theta$ \citep{sutton2018reinforcement}: 

\begin{equation}
    \label{eq:policy-gradient}
    \begin{split}
        J(\pi_\theta) &= \sum_{s \in S} \rho_{\pi_\theta}(s) \; V^{\pi_\theta}(s) \\
                    &= \sum_{s \in S} \rho_{\pi_\theta}(s) \; \sum_{a\in A} Q^{\pi_\theta}(s, a) \ \pi_{\theta}(a|s) 
    \end{split}
\end{equation}

where $\bm{\rho_{\pi_\theta}(s)}$ denotes the stationary probability $\bm{\pi_\theta}$ starting from state $\bm{s_0}$ and transitioning to future states according to the policy $\bm{\pi_\theta}$. To determine the best $\bm{\theta}$ that maximizes the function $\bm{J(\pi_\theta)}$, the gradient $\bm{\nabla_{\theta} J(\theta)}$ is calculated as follows: 

\begin{equation}
    \label{eq:policy-gradient-theorem}
    \begin{split}
        \nabla_{\theta} J(\theta) &= \nabla_{\theta} \biggl( \sum_{s \in S} \rho_{\pi_\theta}(s) \; \sum_{a\in A}  Q^{\pi_\theta}(s, a) \ \pi_{\theta}(a|s) \biggr) \\
                                  &\propto \sum_{s \in S} \mu(s) \; \sum_{a\in A} Q^{\pi_\theta}(s, a) \  \nabla \pi_{\theta}(a|s)\
    \end{split}
\end{equation}

Due to the fact that $\bm{\sum_{s \in  S} \eta(s) = 1}$ and the action space is continuous, Equation \ref{eq:policy-gradient-theorem} can be rewritten as: 

\begin{equation}
    \label{eq:policy-gradient-theorem2}
    \nabla_{\theta} J(\theta) = \mathbb{E}_{s \sim \rho^{\pi_\theta} , a \sim \pi_\theta} \Big [ Q^{\pi_\theta}(s, a) \; \nabla_{\theta} \ln \pi_\theta (a_t|s_t) \Big ]
\end{equation}

Equation \ref{eq:off-policy-gradient-theorem} \citep{silver2014deterministic}, referred to as the off-policy gradient theorem, defines the policy change in relation to the ratio of target policy $\bm{\pi_\theta(a|s)}$ to behavior policy $\bm{\beta(a|s)}$. Take note that the training sample is selected according to the target policy $\bm{s \sim \rho^{\pi_\theta}}$, and the expected return is calculated for the same policy $\bm{\pi_\theta}$, where the training sample adheres to the behavior policy $\bm{\beta(a|s)}$.

\begin{equation}
    \label{eq:off-policy-gradient-theorem}
    \nabla_{\theta} J(\theta) = \mathbb{E}_{s \sim \rho^{\beta} , a \sim \beta} \Big [ \frac{\pi_\theta(a|s)}{\beta_{\theta}(a|s)} Q^{\pi_\theta}(s, a) \; \nabla_{\theta} \ln \pi_\theta (a_t|s_t) \Big ]
\end{equation}

The policy gradient theorem depicted in Equation \ref{eq:policy-gradient} \citep{sutton2000policy} served as the foundation for a variety of other Policy Gradients (PG) algorithms, including REINFORCE, Actor-Critic algorithms, and various multi-agent and distributed actor-critic algorithms.

\subsubsection{Policy-Based Algorithms}
Policy-based algorithms are devoted to improving the gradient descent performance by means of applying different methods such as REINFORCE \citep{williams1992simple}, Trust Region Policy Optimization (TRPO) \citep{schulman2015trust}, Proximal Policy Optimization (PPO) \citep{schulman2017proximal}, Phasic Policy Gradient (PPG) \citep{cobbe2020phasic}, and Stein Variational Policy Gradient (SVPG) \citep{liu2017stein}. 

\paragraph{\textbf{REINFORCE}}
\hfill \break
REINFORCE is a Monte-Carlo policy gradient approach that creates a sample by selecting from an entire episode proportionally to the gradient and updates the policy parameter $\bm{\theta}$ with the step size $\bm{\alpha}$. Given that $\bm{\mathbb{E}_{\pi}[G_t|S_t, A_t] =  Q^{\pi}(s, a)} $, REINFORCE may be defined as follows \citep{sutton2018reinforcement}: 

\begin{equation}
    \label{eq:REINFORCE}
    \nabla_{\theta} J(\theta) = \mathbb{E}_{\pi} \Big [G_t \; \nabla_{\theta} \ln \pi_\theta (A_t|S_t) \Big ]
\end{equation}
The Monte Carlo method has a high variance and, hence, a slow pace of learning. By subtracting the baseline value from the expected return $\bm{G_t}$, REINFORCE decreases variance and accelerates learning while maintaining the bias \citep{sutton2018reinforcement}.

\paragraph{\textbf{Trust Region Policy Optimization (TRPO)}}
\hfill \break
Trust Region Policy Optimization (TRPO) \citep{schulman2015trust} belongs to a category of PG methods: it enhances gradient descent by performing protracted steps inside trust zones specified by a KL-Divergence constraint and updates the policy after each trajectory instead of after each state \citep{almahamid2021reinforcement}. Proximal Policy Optimization (PPO) \citep{schulman2017proximal} may be thought of as an extension of TRPO, where the KL-Divergence constraint is applied as a penalty and the objective is clipped to guarantee that the optimization occurs within a predetermined range \citep{shin2019obstacle}.

\noindent
\textbf{Phasic Policy Gradient (PPG)} \citep{cobbe2020phasic} is an extension of PPO \citep{schulman2017proximal}: it incorporates a recurring auxiliary phase that distills information from the value function into the policy network to enhance the training while maintaining decoupling. 

\paragraph{\textbf{Stein Variational Policy Gradient (SVPG)}}
Stein Variational Policy Gradient (SVPG) \citep{liu2017stein} algorithm updates the policy $\bm{\pi_\theta}$ using Stein variational gradient descent (SVGD) \citep{liu2016stein}, therefore reducing variance and improving convergence. When used in conjunction with REINFORCE and the advantage actor-critic algorithms, SVPG enhances average return and data efficiency \citep{liu2016stein}. 

\subsubsection{Actor-Critic}
The term "Actor-Critic algorithms" refers to a collection of algorithms based on the policy gradients theorem. They are composed of two components: 
\begin{enumerate}
    \item The Actor who is liable of finding the optimal policy $\bm{\pi_\theta}$. 
    \item The Critic who estimates the value function $\bm{Q^{w}(s_t,a_t) \approx Q^{\pi}(s_t,a_t)}$ utilizing a parameterized vector $\bm{w}$ and a policy assessment technique such as temporal-difference learning \citep{silver2014deterministic}. 
\end{enumerate}

The actor can be thought of as a network that is attempting to discover the probability of all possible actions and perform the one with the largest probability, whereas the critic can be thought of as a network that is evaluating the chosen action by assessing the quality of the new state created by the performed action. Numerous algorithms can be classified under the actor-critic category including Deterministic policy gradients (DPG) \citep{silver2014deterministic}, Deep Deterministic Policy Gradient (DDPG) \citep{lillicrap2015continuous}, Twin Delayed Deep Deterministic (TD3) \citep{fujimoto2018addressing}, and many others.
\paragraph{\textbf{Deterministic Policy Gradients (DPG)}}
\hfill \break
Deterministic policy gradients (DPG) algorithms implement a deterministic policy $\bm{\mu(s)}$ instead of a stochastic policy $\bm{\pi(s,a)}$. The deterministic policy is a subset of a stochastic policy in which the target policy objective function is averaged over the state distribution of the behavior policy, as depict in \ref{eq:deterministic-policy-gradient} \citep{silver2014deterministic}.

\begin{equation}
    \label{eq:deterministic-policy-gradient}
    \begin{split}
        J_{\beta}(\mu_{\theta}) &= \int_{S} \rho^{\beta}(s) \ V^{\mu}(s) \ ds \\
                                &= \int_{S} \rho^{\beta}(s) \ Q^{\mu}(s,\mu_{\theta}(s)) \ ds
    \end{split}
\end{equation}

Importance sampling is frequently used in off-policy techniques with a stochastic policy to account for mismatches between behavior and target policies. The deterministic policy gradient eliminates the integral over actions; therefore, the importance sampling can be skipped, resulting in the following gradient \citep{almahamid2021reinforcement}:

\begin{equation}
    \label{eq:deterministic-policy-gradient-theorem}
    \begin{split}
        \nabla_{\theta} J_{\beta}(\mu_{\theta}) &\approx \int_{S} \rho^{\beta}(s) \ \nabla_{\theta} \ \mu_{\theta}(a|s) \ Q^{\mu}(s,\mu_{\theta}(s)) \ ds \\
                                                &= \mathbb{E}_{s \sim \rho^{\beta}} \ \Big [ \nabla_{\theta} \ \mu_{\theta}(s) \nabla_{a} Q^{\mu}(s,a)|_{a=\mu_{\theta}(s)} \Big ]
    \end{split}
\end{equation}

Numerous strategies are employed to enhance DPG; for example, Experience Replay (ER) can be used in conjunction with DPG to increase the stability and efficiency of data \citep{heess2015memory}. Deep Deterministic Policy Gradient (DDPG) \citep{lillicrap2015continuous}, on the other hand, expands DPG by leveraging DQN to operate in continuous action space whereas Twin Delayed Deep Deterministic (TD3) \citep{fujimoto2018addressing} expands on DDPG by utilizing Double DQN to prevent the overestimation of the value function by taking the minimum value between the two critics \citep{fujimoto2018addressing}. 

\paragraph{\textbf{Recurrent Deterministic Policy Gradients  (RDPG)}}
\hfill \break
Wierstra \textit{et al.} \citep{wierstra2010recurrent} applied RNN to Policy Gradient (PG) to build a model-free RL - namely Recurrent Policy Gradient (RPG), for Partially Observable Markov Decision Problem (POMDP), which does not require the agent to have a complete assumption about the environment \citep{wierstra2010recurrent}. RPG applies a method for backpropagating return-weighted characteristic eligibilities through time to approximate a policy gradient for a recurrent neural network \citep{wierstra2010recurrent}.

Recurrent Deterministic Policy Gradient (RDPG) \citep{heess2015memory} implements DPG using RNN and extends the work of RGP \citep{wierstra2010recurrent} to partially observed domains. The RNN with LSTM cells preserves information about past observations over many time steps.

\paragraph{\textbf{Soft Actor-Critic  (SAC)}}
\hfill \break
The objective of Soft Actor-Critic (SAC) is to maximize anticipated reward and the entropy \citep{haarnoja2018soft}. By adding the anticipated entropy of the policy across $\rho_\pi(s_t)$, SAC improves the maximum sum of rewards established by adding the rewards over states transitions $J(\pi) = \sum_{t=1}^{T} \mathbb{E}_{s \sim \rho^{\pi} , a \sim \pi} \Big [ r(s_t,a_t) \Big ]$ \citep{haarnoja2018soft}. Equation \ref{eq:sac-entropy} illustrates an extended entropy goal, in which the temperature parameter $\alpha$ influences the stochasticity of the optimum policy by specifying the importance of the entropy $\mathcal{H}(\pi(.|s_t))$ term to the reward \citep{haarnoja2018soft}. 

\begin{equation}
    \label{eq:sac-entropy}
    J(\pi) = \sum_{t=1}^{T} \mathbb{E}_{s \sim \rho^{\pi} , a \sim \pi} \Big [ r(s_t,a_t) + \alpha \mathcal{H}(\pi(.|s_t)) \Big ]
\end{equation}

By using function approximators and two independent NNs for the actor and critic, SAC estimates a soft Q-function $\bm{Q_\theta(s_t,a_t)}$ parameterized by $\bm{\theta}$, a state value function $\bm{V_\psi(s_t)}$ parameterized by $\bm{\psi}$, and an adjustable policy $\bm{\pi_\phi(a_t|s_t)}$ parameterized by $\bm{\phi}$ \citep{almahamid2021reinforcement}. 

\subsubsection{Multi-Agent and Distributed Actor-Critic}
This group of algorithms includes multi-agent and distributed actor-critic algorithms. They are grouped together as multi-agents can be deployed across several nodes making it a distributed system. 

\paragraph{\textbf{Advantage Actor-Critic}}
\hfill \break
\textbf{Asynchronous Advantage Actor-Critic (A3C)} \citep{mnih2016asynchronous} is a policy gradient algorithm that parallelizes training by using multi-threads, commonly known as workers or agents. Each agent has a local policy $\bm{\pi_\theta(a_t|s_t)}$ and a value function estimate $\bm{V_\theta(s_t)}$. The agent and the same-structured global network asynchronously exchange the parameters in both directions, from agent to the global network and vice-versa. After $t_{max}$ actions or when a final state is reached, the policy and the value function are modified \citep{mnih2016asynchronous}.

\noindent
\textbf{Advantage Actor-Critic (A2C)} \citep{mnih2016asynchronous} is a policy gradient method identical to A3C, except that it includes a coordinator for synchronizing all agents. After all agents complete their work, either by arriving at a final state or by completing $\bm{t_{max}}$ actions, the coordinator updates the policy and value function in both directions between the agents and the global network and vice versa.

Another variant of A3C is \textbf{Actor-Critic with Kronecker-Factored Trust Region (ACKTR)} \citep{wu2017scalable} which uses Kronecker-factored approximation curvature (K-FAC) \citep{martens2015optimizing} to optimize the actor and critic. It improves the computation of natural gradients by efficiently inverting the gradient covariance matrix.

\paragraph{\textbf{Actor-Critic with Experience Replay (ACER)}}
\hfill \break
\noindent
Actor-Critic with Experience Replay (ACER) \citep{wang2016sample} is an off-policy actor-critic algorithm with experience replay that estimates the policy $\pi_\theta(a_t|s_t)$ and the value function $V_{\theta_v}^{\pi}(s_t)$ using a single deep neural network \citep{wang2016sample}. In comparison to A3C, ACER employs a stochastic dueling network and a novel \textit{trust region policy optimization} \citep{wang2016sample}, while improving importance sampling with a bias correction \citep{mnih2016asynchronous}.

ACER applies an improved Retrace algorithm \citep{munos2016safe} by using a truncated importance sampling with bias correction and the value $\bm{Q^{ret}}$ as the target value to train the critic \citep{wang2016sample}. The gradient $\bm{\hat{g}_{t}^{acer}}$ is determined by truncating the importance weights by a constant $\bm{c}$, and subtracting $\bm{V_{\theta_v}(s_t)}$, which reduces variance.

\noindent
\textbf{Retrace-Actor (Reactor)} \citep{gruslys2017reactor} increases sampling and time efficiency by combining contributions from different techniques. 
It employs Distributional Retrace \citep{munos2016safe} to provide multi-step off-policy distributional RL updates while prioritizing replay on transitions \citep{gruslys2017reactor}. Additionally, by taking advantage of action values as a baseline, Reactor improves the trade-off between variance and bias via $\beta$-leave-one-out ($\beta$-LOO) resulting in an improvement of the policy gradient \citep{gruslys2017reactor}. 

\paragraph{\textbf{Multi-Agent Reinforcement Learning (MARL)}}
\hfill \break
\textbf{Distributed Distributional DDPG (D4PG)} \citep{barth2018distributed} adds features such as N-step returns and prioritized experience replay to the distributed settings of DDPG \citep{barth2018distributed}. On the other hand, Multi-Agent DDPG (MADDPG) \citep{lowe2017multi} expands DDPG to coordinate between multiple agents and learn policies while considering each agent's policy \citep{lowe2017multi}. In comparison, Multi-Agent TD3 (MATD3) expands TD3 \citep{ackermann2019reducing} to work with multi-agents using centralized training and decentralized execution while,similarly to TD3, controlling the overestimation bias by employing two centralized critics for each agent.

\noindent
\textbf{Importance Weighted Actor-Learner Architecture (IMPALA)} \citep{espeholt2018impala} is an off-policy algorithm that separates action execution and policy learning. It can be applied using two distinct configurations: 1) a single learner and multiple actors, or 2) multiple synchronous learners and multiple actors.

Using a single learner and several actors, the trajectories generated by the actors are transferred to the learner. Before initiating a new trajectory, the actors are waiting for the learner to update the policy, while the learner simultaneously queues the received trajectories from the actors and constructs the updated policy. Nonetheless, actors may acquire an older version due to their lack of awareness of one another and the lag between the actors and the learner. To address this challenge, IMPALA employs a unique v-trace correction approach that takes into account a truncated importance sampling (IS), defined as the ratio of the learner's policy $\bm{\pi}$ to the actor's present policy $\bm{\mu}$ \citep{almahamid2021reinforcement}. 
Likewise, with multiple synchronous learners, policy parameters are spread across numerous learners who communicate synchronously via a master learner \citep{espeholt2018impala}. 

\noindent
\textbf{Scalable, Efficient Deep-RL (SEED RL)} \citep{espeholt2019seed} provides a scalable architecture that combines IMPALA with R2D2 and can train on millions of frames per second with a lower cost of experiments compared to IMPALA \citep{espeholt2019seed}. SEED moves the inference to the learner while the environments run remotely, introducing a latency issue due to the increased number of remote calls, which is mitigated using a fast communication layer using gRPC.

\section{Problem Formulation and Algorithm Selection} \label{sec:alg-selection}
\begin{figure*}[!t]
    \centering
    \includegraphics[width=.9\linewidth]{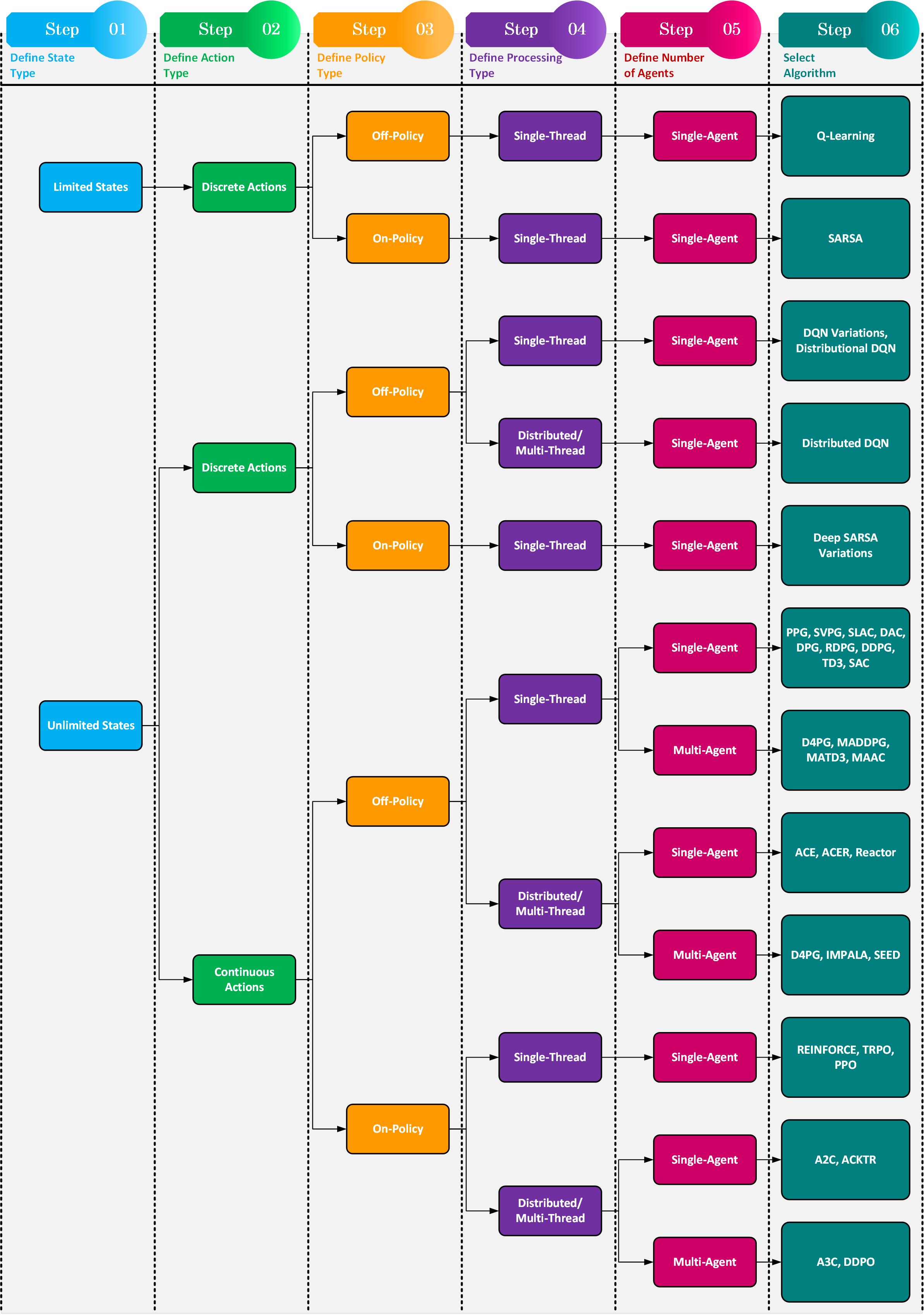}
    \caption{Algorithm Selection Process}
    \label{fig:alg-select-process}
\end{figure*}

The previous section categorized RL algorithms based on state and action types and reviewed the most prominent algorithms. With such a large number of algorithms, it is challenging to select the RL algorithms suitable to tackle the task at hand. Consequently, Figure \ref{fig:alg-select-process} depicts the process of selecting a suitable RL algorithm or a group of RL algorithms through six steps/questions that are answered to guide an informed selection. 

The selection process places a greater emphasis on how the environment and RL objective are formulated than on the RL problem type because the algorithm selection is dependent on the environment and objective formulation. For instance, UAV navigation tasks can employ several sets of algorithms dependent on the desired action type. The six steps, indicated in Figure \ref{fig:alg-select-process}, guide the selection of algorithms: the selected option at each step limits the choices available in the next step based on the available algorithms' characteristics. The steps are as follows:  

\noindent
\textbf{\textbullet \; Step 1 - Define State Type:} When assessing an RL task, it is essential to comprehend the state that can be obtained from the surrounding environment.
For instance, some navigation tasks simplify the environment's states using grid-cell representations \citep{elnaggar2018irl, andrew2018deep, cui2021uav}, where the agent has a limited and predetermined set of states, whereas in other tasks, the environment can have unlimited states \citep{grando2020deep, morad2021embodied, yoon2019hierarchical}. Therefore, this steps involves a decision between limited vs. unlimited states.

\noindent
\textbf{\textbullet \; Step 2 - Define Action Type:} Choosing between discrete and continuous action types limits the number of applicable algorithms.
For instance, discrete actions can be used to move the UAV in pre-specified directions (UP, DOWN, RIGHT, LEFT, etc.), whereas continuous actions, such as the change in pitch, roll, and yaw angles, specify the quantity of the movement using a real number $r\in \mathbb{R}$.

\noindent
\textbf{\textbullet \; Step 3 - Define Policy Type:} As addressed and explained in Subsection \ref{sec:on-policy-vs-off-policy}, RL algorithms can be either off-policy or on-policy algorithms. The policy type selected restricts the alternatives accessible in the subsequent stage. On-policy algorithms converge faster than off-policy algorithms and find a sub-optimal policy, making them a good fit for environments requiring much exploration. Moreover, on-policy algorithms provide stable training since one policy uses learning and data sampling. On the other hand, off-policy algorithms provide an optimal policy and require a good exploration strategy.

The off-policy algorithms' convergence can be improved using techniques such as \textit{prioritized experience replay} and \textit{importance sampling}, making them a good fit for navigation tasks that require finding the optimal path.

\noindent 
\textbf{\textbullet \; Step 4 - Define Processing Type:} While some RL algorithms run in a single thread, others support multi-threading and distributed processing. This steps select the processing type that suits the application needs and the available computational power. 

\noindent 
\textbf{\textbullet \; Step 5 - Define Number of Agents:} This steps specifies the number of agents the application should have. This is needed as some RL algorithms enable MARL, which accelerates learning but requires more computational resources, while other techniques only employ a single agent.

\noindent 
\textbf{\textbullet \; Step 6 - Select the Algorithms:} The last phase of the process results in a collection of algorithms that may be applied to the RL problem at hand. However, the performance of the algorithms is affected by a number of factors and may vary depending on variables such as hyper-parameter settings, reward engineering, and the agent's NN architecture. Consequently, the procedure seeks to reduce the algorithm selection to a group of algorithms rather than a single algorithm. 

While this section presented the process of narrowing down the algorithm for use case, Section 6 provided a description and references to many algorithms to assist in comprehending the distinctions between the algorithms and making an informed selection.

\section{Challenges and Opportunities} \label{sec:challanges-opportunities}
Previous sections demonstrated the diversity of UAV navigation tasks besides the diversity of RL algorithms. Due to such a high number of algorithms, selecting the appropriate algorithms for the task at hand is challenging. Table \ref{tab:rl-algorithms} and discussion in Section \ref{sec:rl-algorithms} provide an overview of the RL algorithms and assist in selecting the RL algorithm for navigation task. Nevertheless, there are still numerous challenges and opportunities in RL for UAV navigation, including:

\noindent{\textbf{Evaluation and benchmarking:}}
Atari 2600 is a home video game console with 57 built-in games that laid the foundation to establish a benchmarking for RL algorithms. The benchmark was established using 57 different games to compare various RL algorithms and set a benchmark baseline against human performance playing the same games. The agent's performance is evaluated and compared to other algorithms using the same benchmark (Atari 2600), or evaluated using various non-navigation environments other than Atari 2600 \citep{andrychowicz2020matters}. The performance of the algorithms using the benchmark might differ when applied to the UAV navigation simulated on a 3D environment or the real world because the games in Atari 2600 can provide a full state of the environment, which means the agent does not need to make assumptions about the state, and MDP can be applied to these problems. Whereas in UAV navigation simulation, the agent knows partial states of the environment (observation), these observations are used to train the agent using POMDP, which results in changing behavior for some of the algorithms.  Furthermore, images processed from the games are 2D images, where the agent in most algorithms tries to learn an optimal policy based on the pattern of the pixels in the image. The same cannot be inferred for images received from the 3D simulators or the real-world images because objects' depth plays a vital role in learning the optimal policy avoiding nearby objects. Therefore, there is a need for new evaluation and benchmarking techniques for RL driven navigation.

\noindent{\textbf{Environment complexity:}}
The tendency to oversimplify the environment and the absence of a standardized benchmarking tools makes it impossible to compare and conclude performances obtained using different algorithms and simulated using various tools and environments. Nevertheless, the UAV needs to perform tasks in different environments and is subject to various conditions, for example:
\begin{itemize}[noitemsep,nolistsep]
    \item Navigating in various environment types such as indoor vs. outdoor.
    \item Considering the changing environment conditions such as wind speed, lighting conditions, and moving objects.
\end{itemize}

Some of the simulation tools discussed in Section \ref{sec:frameworks-sumi-soft}, such as AirSim combined with Unreal Engine, provide different environment types out-of-the-box and are capable of simulating several environmental effects such as changing wind speed and lighting conditions. Still, these complex environments remain to be combined with new benchmarking techniques for improved comparison of RL algorithms for UAV navigation.

\noindent{\textbf{Knowledge transfer:}}
Knowledge transfer imposes another challenge, where the RL agent training in a selected environment does not guarantee similar performance in another environment due to the difference in environments' nature such as different object/obstacles types, background texture, lighting density, and added noise. Most of the existing research focused on applying transfer learning to reduce the training time for the agent in the new environment \citep{yoon2019hierarchical}. However, generalized training methods or other techniques are needed to guarantee a similar performance of the agent in different environments and under various conditions.

\noindent
\textbf{UAVs complexity:} Training UAVs is often accomplished in a 3D virtual environment since UAVs have limited computational resources and power supply, with a typical flight time of 10 to 30 minutes. Reducing the computation time will create possibilities for more complex navigation tasks and increase the flight time since it will reduce energy consumption. Figure 6 shows that only $10\%$ of the investigated research used real drones for navigation training. Therefore, more research is required to focus on energy-aware navigation utilizing low-complexity and efficient RL algorithms while simulating using real drones.

\noindent{\textbf{Algorithm diversity:}}
As seen from Table \ref{tab:rl-algorithms}, many recent and very successful algorithms have not been applied in UAV navigation. As these algorithms have shown great surcease in other domains outperforming the human benchmark, there is a prodigious potential in their application in UAV navigation. The algorithms are expected to gain better generalization on different environments, speed up the training process, and even solve efficiently more complex tasks such as UAVs flocking.

\section{Conclusion} \label{sec:conclusion}
This review deliberates on the application of RL for autonomous UAV navigation. RL uses an intelligent agent to control the UAV movement by processing the states from the environment and moving the UAV in desired directions. The data received from the UAV camera or other sensors such as LiDAR are used to estimate the distance from various objects in the environment and avoid colliding with these objects.

RL algorithms and techniques were used to solve navigation problems such as controlling the UAV while avoiding obstacles, path planning, and flocking. For example, RL is used in single UAV path planning and multi-UAVs flocking to plan path waypoints of the UAV(s) while avoiding obstacles or maintaining flight formation (flocking). Furthermore, this study recognizes various navigation frameworks simulation software used to conduct the experiments along with identifying their use within the reviewed papers.

The review discusses over fifty RL algorithms, explains their contributions and relations, and classifies them according to the application environment and their use in UAV navigation. Furthermore, the study highlights other algorithmic traits such as multi-threading, distributed processing, and multi-agents, followed by a systematic process that aims to assist in finding the set of applicable algorithms.

The study observes that the research community tends to experiment with a specific set of algorithms: Q-learning, DQN, Double DQN, DDPG, PPO, although some recent algorithms show more promising results than the mentioned algorithms such as agent57. To the best of the authors' knowledge, this study is the first systematic review identifying a large number of RL algorithms while focusing on their application in autonomous UAV navigation. 

Analysis of the current RL algorithms and their use in UAV navigation identified the following challenges and opportunities: the need for navigation-focused evaluation and benchmarking techniques, the necessity to work with more complex environments, the need to examine knowledge transfer, the complexity of UAVs, and the necessity to evaluate state-of-the-art RL algorithms on navigation tasks.

\appendix
\section{Acronyms}
\begin{itemize}[noitemsep,nolistsep]
\item \textbf{A2C	:}	Advantage Actor-Critic
\item \textbf{A3C	:}	Asynchronous Advantage Actor-Critic
\item \textbf{AC	:}	Actor-Critic
\item \textbf{ACE	:}	Actor Ensemble
\item \textbf{ACER	:}	Actor-Critic with Experience Replay
\item \textbf{ACKTR	:}	Actor-Critic using Kronecker-Factored Trust Region
\item \textbf{Agent57	:}	Agent57
\item \textbf{Ape-X DPG	:}	Ape-X Deterministic Policy Gradients
\item \textbf{Ape-X DQN	:}	Ape-X Deep Q-Networks
\item \textbf{AS	:}	Autonomous Systems
\item \textbf{C51-DQN	:}	Categorical Deep Q-Networks
\item \textbf{CNN	:}	Recurrent Neural Network
\item \textbf{D4PG	:}	Distributed Distributional DDPG
\item \textbf{DAC	:}	Double Actor-Critic
\item \textbf{DD-DQN	:}	Double Dueling Deep Q-Networks
\item \textbf{DD-DRQN	:}	Double Dueling Deep Recurrent Q-Networks
\item \textbf{DDPG	:}	Deep Deterministic Policy Gradient
\item \textbf{Double DQN	:}	Double Deep Q-Networks
\item \textbf{DPG	:}	Deterministic Policy Gradients
\item \textbf{DPPO	:}	Distributed Proximal Policy Optimization
\item \textbf{DQN	:}	Deep Q-Networks
\item \textbf{DRL	:}	Deep Reinforcement Learning
\item \textbf{DRQN	:}	Deep Recurrent Q-Networks
\item \textbf{Dueling DQN	:}	Dueling Deep Q-Networks
\item \textbf{DVS	:}	Dynamic Vision Sensor
\item \textbf{eNVM	:}	embedded Non-Volatile Memory
\item \textbf{FOV	:}	Field Of View
\item \textbf{FQF	:}	Fully parameterized Quantile Function
\item \textbf{GPS	:}	Global Positioning System
\item \textbf{IMPALA	:}	Importance Weighted Actor-Learner Architecture
\item \textbf{IMU	:}	Inertial Measurement Unit
\item \textbf{IQN	:}	Implicit Quantile Networks
\item \textbf{K-FAC	:}	Kronecker-factored approximation curvature
\item \textbf{KL	:}	Kullback-Leibler
\item \textbf{LSTM	:}	Long-Short Term Memory
\item \textbf{MAAC	:}	Multi-Actor-Attention-Critic
\item \textbf{MADDPG	:}	Multi-Agent DDPG
\item \textbf{MARL	:}	Multi-Agent Reinforcement Learning
\item \textbf{MATD3	:}	Multi-Agent Twin Delayed Deep Deterministic
\item \textbf{MATD3	:}	Multi-Agent TD3
\item \textbf{MDP	:}	Markov Decision Problem
\item \textbf{NGA	:}	Never Give Up
\item \textbf{Noisy DQN	:}	Noisy Deep Q-Networks
\item \textbf{PAAC	:}	Parallel Advantage Actor-Critic
\item \textbf{PER	:}	Prioritized Experience Replay
\item \textbf{PG	:}	Policy Gradients
\item \textbf{POMDP	:}	Partially Observable Markov Decision Problem
\item \textbf{PPG	:}	Phasic Policy Gradient
\item \textbf{PPO	:}	Proximal Policy Optimization
\item \textbf{QR-DQN	:}	Quantile Regression Deep Q-Networks
\item \textbf{R2D2	:}	Recurrent Replay Distributed Deep Q-Networks
\item \textbf{Rainbow DQN	:}	Rainbow Deep Q-Networks
\item \textbf{RDPG	:}	Recurrent Deterministic Policy Gradients
\item \textbf{Reactor	:}	Retrace-Actor
\item \textbf{REINFORCE	:}	\textbf{RE}ward \textbf{I}ncrement $=$ \textbf{N}onnegative \textbf{F}actor $\times$ \textbf{O}ffset \textbf{R}einforcement $\times$ \textbf{C}haracteristic \textbf{E}ligibility
\item \textbf{RL	:}	Reinforcement Learning
\item \textbf{RND	:}	Random Network Distillation
\item \textbf{RNN	:}	Recurrent Neural Network
\item \textbf{ROS	:}	Robot Operating System
\item \textbf{SAC	:}	Soft Actor-Critic
\item \textbf{SARSA	:}	State-Action-Reward-State-Action
\item \textbf{SEED RL	:}	Scalable, Efficient Deep-RL
\item \textbf{SLAC	:}	Stochastic Latent Actor-Critic
\item \textbf{SRAM	:}	Static Random Access Memory
\item \textbf{SVPG	:}	Stein Variational Policy Gradient
\item \textbf{TD	:}	Temporal Difference
\item \textbf{TD3	:}	Twin Delayed Deep Deterministic
\item \textbf{TRPO	:}	Trust Region Policy Optimization
\item \textbf{UAV	:}	Unmanned Aerial Vehicle
\item \textbf{UBC	:}	Upper Confidence Bound
\item \textbf{UGV	:}	Unmanned Ground Vehicle
\item \textbf{UVFA	:}	Universal Value Function Approximator
\end{itemize}

\printcredits

\section*{Acknowledgments}

\noindent 
This research has been supported by NSERC under grant RGPIN-2018-06222

\bibliographystyle{elsarticle-num}
\bibliography{RL_UAV_Survey.bib}

\begin{thebibliography}{100}
\expandafter\ifx\csname url\endcsname\relax
  \def\url#1{\texttt{#1}}\fi
\expandafter\ifx\csname urlprefix\endcsname\relax\def\urlprefix{URL }\fi
\expandafter\ifx\csname href\endcsname\relax
  \def\href#1#2{#2} \def\path#1{#1}\fi

\bibitem{lu2018survey}
Y.~Lu, Z.~Xue, G.-S. Xia, L.~Zhang, {A survey on vision-based UAV navigation},
  Taylor \& Francis Geo-spatial information science 21~(1) (2018) 21--32.
\newblock \href {https://doi.org/10.1080/10095020.2017.1420509}
  {\path{doi:10.1080/10095020.2017.1420509}}.

\bibitem{zeng2020survey}
F.~Zeng, C.~Wang, S.~S. Ge, {A survey on visual navigation for artificial
  agents with deep reinforcement learning}, IEEE Access 8 (2020)
  135426--135442.

\bibitem{azoulay2021machine}
R.~Azoulay, Y.~Haddad, S.~Reches, {Machine Learning Methods for UAV Flocks
  Management-A Survey}, IEEE Access 9 (2021) 139146--139175.

\bibitem{aggarwal2020path}
S.~Aggarwal, N.~Kumar, {Path planning techniques for unmanned aerial vehicles:
  A review, solutions, and challenges}, Computer Communications 149 (2020)
  270--299.

\bibitem{lou2016adaptive}
W.~Lou, X.~Guo, {Adaptive trajectory tracking control using reinforcement
  learning for quadrotor}, SAGE International Journal of Advanced Robotic
  Systems 13~(1) (2016).
\newblock \href {https://doi.org/10.5772/62128} {\path{doi:10.5772/62128}}.

\bibitem{guerra2021networks}
A.~Guerra, F.~Guidi, D.~Dardari, P.~M. Djuri{\'c}, {Networks of UAVs of
  low-complexity for time-critical localization}, arXiv preprint
  arXiv:2108.13181 (2021).

\bibitem{guerra2021real}
A.~Guerra, F.~Guidi, D.~Dardari, P.~M. Djuri{\'c}, {Real-Time Learning for THZ
  Radar Mapping and UAV Control}, in: IEEE International Conference on
  Autonomous Systems, 2021, pp. 1--5.
\newblock \href {https://doi.org/10.1109/ICAS49788.2021.9551141}
  {\path{doi:10.1109/ICAS49788.2021.9551141}}.

\bibitem{guerra2020dynamic}
A.~Guerra, D.~Dardari, P.~M. Djuri{\'c}, {Dynamic radar network of UAVs: A
  joint navigation and tracking approach}, IEEE Access 8 (2020) 116454--116469.

\bibitem{liu2020distributed}
Y.~Liu, Y.~Wang, J.~Wang, Y.~Shen, {Distributed 3D relative localization of
  UAVs}, IEEE Transactions on Vehicular Technology 69 (2020) 11756--11770.

\bibitem{zhang2020self}
S.~Zhang, R.~P{\"o}hlmann, T.~Wiedemann, A.~Dammann, H.~Wymeersch, P.~A.
  Hoeher, {Self-aware swarm navigation in autonomous exploration missions},
  Proceedings of the IEEE 108~(7) (2020) 1168--1195.

\bibitem{almahamid2021reinforcement}
F.~AlMahamid, K.~Grolinger, {Reinforcement Learning Algorithms: An Overview and
  Classification}, in: IEEE Canadian Conference on Electrical and Computer
  Engineering, 2021, pp. 1--7.

\bibitem{sutton2018reinforcement}
R.~S. Sutton, A.~G. Barto, {Reinforcement Learning: An Introduction}, MIT
  Press, 2018.

\bibitem{lin1992self}
L.-J. Lin, {Self-improving reactive agents based on reinforcement learning,
  planning and teaching}, Springer Machine learning 8~(3-4) (1992) 293--321.

\bibitem{schaul2015prioritized}
T.~Schaul, J.~Quan, I.~Antonoglou, D.~Silver, {Prioritized experience replay},
  arXiv:1511.05952 (2015).

\bibitem{silver2014deterministic}
D.~Silver, G.~Lever, N.~Heess, T.~Degris, D.~Wierstra, M.~Riedmiller,
  {Deterministic Policy Gradient Algorithms}, in: PMLR International conference
  on machine learning, 2014, pp. 387--395.

\bibitem{zhou2020efficient}
S.~Zhou, B.~Li, C.~Ding, L.~Lu, C.~Ding, {An Efficient Deep Reinforcement
  Learning Framework for UAVs}, in: IEEE International Symposium on Quality
  Electronic Design, 2020, pp. 323--328.

\bibitem{karthik2020reinforcement}
P.~Karthik, K.~Kumar, V.~Fernandes, K.~Arya, {Reinforcement Learning for
  Altitude Hold and Path Planning in a Quadcopter}, in: IEEE International
  Conference on Control, Automation and Robotics, 2020, pp. 463--467.

\bibitem{deshpande2020developmental}
A.~M. Deshpande, R.~Kumar, A.~A. Minai, M.~Kumar, {Developmental reinforcement
  learning of control policy of a quadcopter UAV with thrust vectoring rotors},
  in: ASME Dynamic Systems and Control Conference, Vol. 84287, 2020, p.
  V002T36A011.
\newblock \href {https://doi.org/10.1115/DSCC2020-3319}
  {\path{doi:10.1115/DSCC2020-3319}}.

\bibitem{camci2019learning}
E.~Camci, E.~Kayacan, {Learning motion primitives for planning swift maneuvers
  of quadrotor}, Springer Autonomous Robots 43~(7) (2019) 1733--1745.

\bibitem{li2019optimal}
S.~Li, P.~Durdevic, Z.~Yang, {Optimal Tracking Control Based on Integral
  Reinforcement Learning for An Underactuated Drone}, Elsevier
  IFAC-PapersOnLine 52~(8) (2019) 194--199.

\bibitem{greatwood2019reinforcement}
C.~Greatwood, A.~G. Richards, {Reinforcement learning and model predictive
  control for robust embedded quadrotor guidance and control}, Springer
  Autonomous Robots 43~(7) (2019) 1681--1693.

\bibitem{koch2019reinforcement}
W.~Koch, R.~Mancuso, R.~West, A.~Bestavros, {Reinforcement learning for UAV
  attitude control}, ACM Transactions on Cyber-Physical Systems 3~(2) (2019)
  1--21.

\bibitem{salvatore2020neuro}
N.~Salvatore, S.~Mian, C.~Abidi, A.~D. George, {A Neuro-Inspired Approach to
  Intelligent Collision Avoidance and Navigation}, in: IEEE Digital Avionics
  Systems Conference, 2020, pp. 1--9.

\bibitem{bouhamed2020uav}
O.~Bouhamed, H.~Ghazzai, H.~Besbes, Y.~Massoud, {A UAV-Assisted Data Collection
  for Wireless Sensor Networks: Autonomous Navigation and Scheduling}, IEEE
  Access 8 (2020) 110446--110460.

\bibitem{huang2019autonomous}
H.~Huang, J.~Gu, Q.~Wang, Y.~Zhuang, {An Autonomous UAV Navigation System for
  Unknown Flight Environment}, in: IEEE International Conference on Mobile
  Ad-Hoc and Sensor Networks, 2019, pp. 63--68.

\bibitem{shin2019automatic}
S.-Y. Shin, Y.-W. Kang, Y.-G. Kim, {Automatic Drone Navigation in Realistic 3D
  Landscapes using Deep Reinforcement Learning}, in: IEEE International
  Conference on Control, Decision and Information Technologies, 2019, pp.
  1072--1077.

\bibitem{wang2017autonomous}
C.~Wang, J.~Wang, X.~Zhang, X.~Zhang, {Autonomous Navigation of UAV in
  Large-Scale Unknown Complex Environment with Deep Reinforcement Learning},
  in: IEEE Global Conference on Signal and Information Processing, 2017, pp.
  858--862.

\bibitem{wang2019autonomous}
C.~Wang, J.~Wang, Y.~Shen, X.~Zhang, {Autonomous Navigation of UAVs in
  Large-Scale Complex Environments: A Deep Reinforcement Learning Approach},
  IEEE Transactions on Vehicular Technology 68~(3) (2019) 2124--2136.

\bibitem{anwar2020autonomous}
A.~Anwar, A.~Raychowdhury, {Autonomous Navigation via Deep Reinforcement
  Learning for Resource Constraint Edge Nodes using Transfer Learning}, IEEE
  Access 8 (2020) 26549--26560.

\bibitem{bouhamed2020autonomous}
O.~Bouhamed, H.~Ghazzai, H.~Besbes, Y.~Massoud, {Autonomous UAV Navigation: A
  DDPG-Based Deep Reinforcement Learning Approach}, in: IEEE International
  Symposium on Circuits and Systems, 2020, pp. 1--5.

\bibitem{yang2020autonomous}
Y.~Yang, K.~Zhang, D.~Liu, H.~Song, {Autonomous UAV Navigation in Dynamic
  Environments with Double Deep Q-Networks}, in: IEEE Digital Avionics Systems
  Conference, 2020, pp. 1--7.

\bibitem{li2019autonomous}
Y.~Li, M.~Li, A.~Sanyal, Y.~Wang, Q.~Qiu, {Autonomous UAV with Learned
  Trajectory Generation and Control}, in: IEEE International Workshop on Signal
  Processing Systems, 2019, pp. 115--120.

\bibitem{chen2020collision}
Y.~Chen, N.~Gonz{\'a}lez-Prelcic, R.~W. Heath, {Collision-free UAV navigation
  with a monocular camera using deep reinforcement learning}, in: IEEE
  International Workshop on Machine Learning for Signal Processing, 2020, pp.
  1--6.

\bibitem{grando2020deep}
R.~B. Grando, J.~C. de~Jesus, P.~L. Drews-Jr, {Deep Reinforcement Learning for
  Mapless Navigation of Unmanned Aerial Vehicles}, in: IEEE Latin American
  Robotics Symposium, Brazilian Symposium on Robotics and Workshop on Robotics
  in Education, 2020, pp. 1--6.

\bibitem{camci2020deep}
E.~Camci, D.~Campolo, E.~Kayacan, {Deep Reinforcement Learning for Motion
  Planning of Quadrotors Using Raw Depth Images}, Learning (RL) 10 (2020).

\bibitem{wang2020deep}
C.~Wang, J.~Wang, J.~Wang, X.~Zhang, {Deep-Reinforcement-Learning-Based
  Autonomous UAV Navigation With Sparse Rewards}, IEEE Internet of Things
  Journal 7~(7) (2020) 6180--6190.

\bibitem{cetin2019drone}
E.~Cetin, C.~Barrado, G.~Mu{\~n}oz, M.~Macias, E.~Pastor, {Drone navigation and
  avoidance of obstacles through deep reinforcement learning}, in: IEEE Digital
  Avionics Systems Conference, 2019, pp. 1--7.

\bibitem{morad2021embodied}
S.~D. Morad, R.~Mecca, R.~P. Poudel, S.~Liwicki, R.~Cipolla, {Embodied Visual
  Navigation with Automatic Curriculum Learning in Real Environments}, IEEE
  Robotics and Automation Letters 6~(2) (2021) 683--690.

\bibitem{yan2020flocking}
P.~Yan, C.~Bai, H.~Zheng, J.~Guo, {Flocking Control of UAV Swarms with Deep
  Reinforcement Learning Approach}, in: IEEE International Conference on
  Unmanned Systems, 2020, pp. 592--599.

\bibitem{yoon2019hierarchical}
I.~Yoon, M.~A. Anwar, R.~V. Joshi, T.~Rakshit, A.~Raychowdhury, {Hierarchical
  memory system with STT-MRAM and SRAM to support transfer and real-time
  reinforcement learning in autonomous drones}, IEEE Journal on Emerging and
  Selected Topics in Circuits and Systems 9~(3) (2019) 485--497.

\bibitem{williams2017information}
G.~Williams, N.~Wagener, B.~Goldfain, P.~Drews, J.~M. Rehg, B.~Boots, E.~A.
  Theodorou, {Information theoretic MPC for model-based reinforcement
  learning}, in: IEEE International Conference on Robotics and Automation,
  2017, pp. 1714--1721.

\bibitem{he2020integrated}
L.~He, N.~Aouf, J.~F. Whidborne, B.~Song, {Integrated moment-based LGMD and
  deep reinforcement learning for UAV obstacle avoidance}, in: IEEE
  International Conference on Robotics and Automation, 2020, pp. 7491--7497.

\bibitem{singla2019memory}
A.~Singla, S.~Padakandla, S.~Bhatnagar, {Memory-based deep reinforcement
  learning for obstacle avoidance in UAV with limited environment knowledge},
  IEEE Transactions on Intelligent Transportation Systems (2019).

\bibitem{wu2018navigating}
T.-C. Wu, S.-Y. Tseng, C.-F. Lai, C.-Y. Ho, Y.-H. Lai, {Navigating assistance
  system for quadcopter with deep reinforcement learning}, in: IEEE
  International Cognitive Cities Conference, 2018, pp. 16--19.

\bibitem{anwar2018navren}
M.~A. Anwar, A.~Raychowdhury, {NavREn-Rl: Learning to fly in real environment
  via end-to-end deep reinforcement learning using monocular images}, in: IEEE
  International Conference on Mechatronics and Machine Vision in Practice,
  2018, pp. 1--6.

\bibitem{zhou2018neural}
B.~Zhou, W.~Wang, Z.~Wang, B.~Ding, {Neural Q-learning algorithm based UAV
  obstacle avoidance}, in: IEEE CSAA Guidance, Navigation and Control
  Conference, 2018, pp. 1--6.

\bibitem{yijing2017q}
Z.~Yijing, Z.~Zheng, Z.~Xiaoyi, L.~Yang, {Q-learning algorithm based UAV path
  learning and obstacle avoidance approach}, in: IEEE Chinese Control
  Conference, 2017, pp. 3397--3402.

\bibitem{villanueva2019deep}
A.~Villanueva, A.~Fajardo, {Deep Reinforcement Learning with Noise Injection
  for UAV Path Planning}, in: IEEE International Conference on Engineering
  Technologies and Applied Sciences, 2019, pp. 1--6.

\bibitem{walvekar2019vision}
A.~Walvekar, Y.~Goel, A.~Jain, S.~Chakrabarty, A.~Kumar, {Vision based
  autonomous navigation of quadcopter using reinforcement learning}, in: IEEE
  International Conference on Automation, Electronics and Electrical
  Engineering, 2019, pp. 160--165.

\bibitem{zhou2019vision}
B.~Zhou, W.~Wang, Z.~Liu, J.~Wang, {Vision-based Navigation of UAV with
  Continuous Action Space Using Deep Reinforcement Learning}, in: IEEE Chinese
  Control And Decision Conference, 2019, pp. 5030--5035.

\bibitem{hasanzade2021dynamically}
M.~Hasanzade, E.~Koyuncu, {A Dynamically Feasible Fast Replanning Strategy with
  Deep Reinforcement Learning}, Springer Journal of Intelligent \& Robotic
  Systems 101~(1) (2021) 1--17.

\bibitem{munoz2019deep}
G.~Mu{\~n}oz, C.~Barrado, E.~{\c{C}}etin, E.~Salami, {Deep reinforcement
  learning for drone delivery}, MDPI Drones 3~(3) (2019).
\newblock \href {https://doi.org/10.3390/drones3030072}
  {\path{doi:10.3390/drones3030072}}.

\bibitem{hodge2021deep}
V.~J. Hodge, R.~Hawkins, R.~Alexander, {Deep reinforcement learning for drone
  navigation using sensor data}, Springer Neural Computing and Applications
  33~(6) (2021) 2015--2033.

\bibitem{doukhi2021deep}
O.~Doukhi, D.-J. Lee, {Deep Reinforcement Learning for End-to-End Local Motion
  Planning of Autonomous Aerial Robots in Unknown Outdoor Environments:
  Real-Time Flight Experiments}, MDPI Sensors 21~(7) (2021) 2534.
\newblock \href {https://doi.org/10.3390/s21072534}
  {\path{doi:10.3390/s21072534}}.

\bibitem{bakale2020indoor}
V.~A. Bakale, Y.~K. VS, V.~C. Roodagi, Y.~N. Kulkarni, M.~S. Patil,
  S.~Chickerur, {Indoor Navigation with Deep Reinforcement Learning}, in: IEEE
  International Conference on Inventive Computation Technologies, 2020, pp.
  660--665.

\bibitem{maxey2019navigation}
C.~J. Maxey, E.~J. Shamwell, {Navigation and collision avoidance with human
  augmented supervisory training and fine tuning via reinforcement learning},
  in: SPIE Micro-and Nanotechnology Sensors, Systems, and Applications XI, Vol.
  10982, 2019, pp. 325 -- 334.
\newblock \href {https://doi.org/10.1117/12.2518551}
  {\path{doi:10.1117/12.2518551}}.

\bibitem{zhao2021reinforcement}
Y.~Zhao, J.~Guo, C.~Bai, H.~Zheng, {Reinforcement Learning-Based Collision
  Avoidance Guidance Algorithm for Fixed-Wing UAVs}, Hindawi Complexity 2021
  (2021).
\newblock \href {https://doi.org/10.1155/2021/8818013}
  {\path{doi:10.1155/2021/8818013}}.

\bibitem{tong2021uav}
G.~Tong, N.~Jiang, L.~Biyue, Z.~Xi, W.~Ya, D.~Wenbo, {UAV navigation in high
  dynamic environments: A deep reinforcement learning approach}, Elsevier
  Chinese Journal of Aeronautics 34~(2) (2021) 479--489.

\bibitem{bouhamed2020ddpg}
O.~Bouhamed, X.~Wan, H.~Ghazzai, Y.~Massoud, {A DDPG-Based Approach for
  Energy-aware UAV Navigation in Obstacle-constrained Environment}, in: IEEE
  World Forum on Internet of Things, 2020, pp. 1--6.

\bibitem{walker2019deep}
O.~Walker, F.~Vanegas, F.~Gonzalez, S.~Koenig, {A Deep Reinforcement Learning
  Framework for UAV Navigation in Indoor Environments}, in: IEEE Aerospace
  Conference, 2019, pp. 1--14.

\bibitem{bouhamed2020generic}
O.~Bouhamed, H.~Ghazzai, H.~Besbes, Y.~Massoud, {A Generic Spatiotemporal
  Scheduling for Autonomous UAVs: A Reinforcement Learning-Based Approach},
  IEEE Open Journal of Vehicular Technology 1 (2020) 93--106.

\bibitem{zhang2020iadrl}
J.~Zhang, Z.~Yu, S.~Mao, S.~C. Periaswamy, J.~Patton, X.~Xia, {IADRL: Imitation
  augmented deep reinforcement learning enabled UGV-UAV coalition for tasking
  in complex environments}, IEEE Access 8 (2020) 102335--102347.

\bibitem{yu2019navigation}
X.~Yu, Y.~Wu, X.-M. Sun, {A Navigation Scheme for a Random Maze Using
  Reinforcement Learning with Quadrotor Vision}, in: IEEE European Control
  Conference, 2019, pp. 518--523.

\bibitem{li2018path}
H.~Li, S.~Wu, P.~Xie, Z.~Qin, B.~Zhang, {A Path Planning for One UAV Based on
  Geometric Algorithm}, in: IEEE CSAA Guidance, Navigation and Control
  Conference, 2018, pp. 1--5.

\bibitem{sacharny2019optimal}
D.~Sacharny, T.~C. Henderson, {Optimal Policies in Complex Large-scale UAS
  Traffic Management}, in: IEEE International Conference on Industrial Cyber
  Physical Systems, 2019, pp. 352--357.

\bibitem{camci2019planning}
E.~Camci, E.~Kayacan, {Planning swift maneuvers of quadcopter using motion
  primitives explored by reinforcement learning}, in: IEEE American Control
  Conference, 2019, pp. 279--285.

\bibitem{guerra2020reinforcement}
A.~Guerra, F.~Guidi, D.~Dardari, P.~M. Djuric, {Reinforcement learning for UAV
  autonomous navigation, mapping and target detection}, in: ION Position,
  Location and Navigation Symposium, 2020, pp. 1004--1013.

\bibitem{cui2021uav}
Z.~Cui, Y.~Wang, {UAV Path Planning Based on Multi-Layer Reinforcement Learning
  Technique}, IEEE Access 9 (2021) 59486--59497.

\bibitem{wang2021pretrained}
Z.~Wang, H.~Li, Z.~Wu, H.~Wu, A pretrained proximal policy optimization
  algorithm with reward shaping for aircraft guidance to a moving destination
  in three-dimensional continuous space, SAGE International Journal of Advanced
  Robotic Systems 18~(1) (2021).
\newblock \href {https://doi.org/10.1177/1729881421989546}
  {\path{doi:10.1177/1729881421989546}}.

\bibitem{eslamiat2019autonomous}
H.~Eslamiat, Y.~Li, N.~Wang, A.~K. Sanyal, Q.~Qiu, {Autonomous waypoint
  planning, optimal trajectory generation and nonlinear tracking control for
  multi-rotor UAVs}, in: IEEE European Control Conference, 2019, pp.
  2695--2700.

\bibitem{imanberdiyev2016autonomous}
N.~Imanberdiyev, C.~Fu, E.~Kayacan, I.-M. Chen, {Autonomous Navigation of UAV
  by using Real-Time Model-Based Reinforcement Learning}, in: IEEE
  International conference on control, automation, robotics and vision, 2016,
  pp. 1--6.

\bibitem{abedin2020data}
S.~F. Abedin, M.~S. Munir, N.~H. Tran, Z.~Han, C.~S. Hong, {Data freshness and
  energy-efficient UAV navigation optimization: A deep reinforcement learning
  approach}, IEEE Transactions on Intelligent Transportation Systems (2020).

\bibitem{andrew2018deep}
W.~Andrew, C.~Greatwood, T.~Burghardt, {Deep learning for exploration and
  recovery of uncharted and dynamic targets from UAV-like vision}, in: IEEE RSJ
  International Conference on Intelligent Robots and Systems, 2018, pp.
  1124--1131.

\bibitem{pham2018reinforcement}
H.~X. Pham, H.~M. La, D.~Feil-Seifer, L.~Van~Nguyen, {Reinforcement learning
  for autonomous UAV navigation using function approximation}, in: IEEE
  International Symposium on Safety, Security, and Rescue Robotics, 2018, pp.
  1--6.

\bibitem{kulkarni2020uav}
S.~Kulkarni, V.~Chaphekar, M.~M.~U. Chowdhury, F.~Erden, I.~Guvenc, {UAV aided
  search and rescue operation using reinforcement learning}, in: IEEE
  SoutheastCon, Vol.~2, 2020, pp. 1--8.

\bibitem{peake2020wilderness}
A.~Peake, J.~McCalmon, Y.~Zhang, B.~Raiford, S.~Alqahtani, {Wilderness search
  and rescue missions using deep reinforcement learning}, in: IEEE
  International Symposium on Safety, Security, and Rescue Robotics, 2020, pp.
  102--107.

\bibitem{akhloufi2019drones}
M.~A. Akhloufi, S.~Arola, A.~Bonnet, {Drones chasing drones: Reinforcement
  learning and deep search area proposal}, MDPI Drones 3~(3) (2019).
\newblock \href {https://doi.org/10.3390/drones3030058}
  {\path{doi:10.3390/drones3030058}}.

\bibitem{polvara2018toward}
R.~Polvara, M.~Patacchiola, S.~Sharma, J.~Wan, A.~Manning, R.~Sutton,
  A.~Cangelosi, {Toward end-to-end control for UAV autonomous landing via deep
  reinforcement learning}, in: IEEE International conference on unmanned
  aircraft systems, 2018, pp. 115--123.

\bibitem{polvara2019autonomous}
R.~Polvara, S.~Sharma, J.~Wan, A.~Manning, R.~Sutton, {Autonomous Vehicular
  Landings on the Deck of an Unmanned Surface Vehicle using Deep Reinforcement
  Learning}, Cambridge Core Robotica 37~(11) (2019) 1867–1882.
\newblock \href {https://doi.org/10.1017/S0263574719000316}
  {\path{doi:10.1017/S0263574719000316}}.

\bibitem{lee2018vision}
S.~Lee, T.~Shim, S.~Kim, J.~Park, K.~Hong, H.~Bang, {Vision-based autonomous
  landing of a multi-copter unmanned aerial vehicle using reinforcement
  learning}, in: IEEE International Conference on Unmanned Aircraft Systems,
  2018, pp. 108--114.

\bibitem{wang2018deep}
C.~Wang, J.~Wang, X.~Zhang, {A Deep Reinforcement Learning Approach to Flocking
  and Navigation of UAVs in Large-Scale Complex Environments}, in: IEEE Global
  Conference on Signal and Information Processing, 2018, pp. 1228--1232.

\bibitem{lee2020autonomous}
G.~T. Lee, C.~O. Kim, {Autonomous Control of Combat Unmanned Aerial Vehicles to
  Evade Surface-to-Air Missiles Using Deep Reinforcement Learning}, IEEE Access
  8 (2020) 226724--226736.

\bibitem{madridano2021software}
{\'A}.~Madridano, A.~Al-Kaff, P.~Flores, D.~Mart{\'\i}n, A.~de~la Escalera,
  {Software Architecture for Autonomous and Coordinated Navigation of UAV
  Swarms in Forest and Urban Firefighting}, MDPI Applied Sciences 11~(3)
  (2021).
\newblock \href {https://doi.org/10.3390/app11031258}
  {\path{doi:10.3390/app11031258}}.

\bibitem{wang2020two}
D.~Wang, T.~Fan, T.~Han, J.~Pan, {A Two-Stage Reinforcement Learning Approach
  for Multi-UAV Collision Avoidance Under Imperfect Sensing}, IEEE Robotics and
  Automation Letters 5~(2) (2020) 3098--3105.

\bibitem{moon2021deep}
J.~Moon, S.~Papaioannou, C.~Laoudias, P.~Kolios, S.~Kim, {Deep Reinforcement
  Learning Multi-UAV Trajectory Control for Target Tracking}, IEEE Internet of
  Things Journal (2021).

\bibitem{liu2019distributed}
C.~H. Liu, X.~Ma, X.~Gao, J.~Tang, {Distributed energy-efficient multi-UAV
  navigation for long-term communication coverage by deep reinforcement
  learning}, IEEE Transactions on Mobile Computing 19~(6) (2019) 1274--1285.

\bibitem{omi2021introduction}
S.~Omi, H.-S. Shin, A.~Tsourdos, J.~Espeland, A.~Buchi, {Introduction to UAV
  swarm utilization for communication on the move terminals tracking evaluation
  with reinforcement learning technique}, in: IEEE European Conference on
  Antennas and Propagation, 2021, pp. 1--5.

\bibitem{viseras2021wildfire}
A.~Viseras, M.~Meissner, J.~Marchal, {Wildfire Front Monitoring with Multiple
  UAVs using Deep Q-Learning}, IEEE Access (2021).

\bibitem{bonnet2019uav}
A.~Bonnet, M.~A. Akhloufi, {UAV pursuit using reinforcement learning}, in: SPIE
  Unmanned Systems Technology XXI, Vol. 11021, International Society for Optics
  and Photonics, 2019, pp. 51 -- 58.
\newblock \href {https://doi.org/10.1117/12.2520310}
  {\path{doi:10.1117/12.2520310}}.

\bibitem{fan2020prioritized}
S.~Fan, G.~Song, B.~Yang, X.~Jiang, {Prioritized Experience Replay in
  Multi-Actor-Attention-Critic for Reinforcement Learning}, IOPscience Journal
  of Physics: Conference Series 1631 (2020).
\newblock \href {https://doi.org/10.1088/1742-6596/1631/1/012040}
  {\path{doi:10.1088/1742-6596/1631/1/012040}}.

\bibitem{majd2018integrating}
A.~Majd, A.~Ashraf, E.~Troubitsyna, M.~Daneshtalab, {Integrating learning,
  optimization, and prediction for efficient navigation of swarms of drones},
  in: IEEE Euromicro International Conference on Parallel, Distributed and
  Network-based Processing, 2018, pp. 101--108.

\bibitem{chapman2016dronetypes}
A.~Chapman, {Drone Types: Multi-Rotor vs Fixed-Wing vs Single Rotor vs Hybrid
  VTOL}, \url{https://www.auav.com.au/articles/drone-types/}, (Accessed:
  01.11.2021) (2016).

\bibitem{elnaggar2018irl}
M.~Elnaggar, N.~Bezzo, {An IRL Approach for Cyber-Physical Attack Intention
  Prediction and Recovery}, in: IEEE American Control Conference, 2018, pp.
  222--227.

\bibitem{escobar2018r}
H.~D. Escobar-Alvarez, N.~Johnson, T.~Hebble, K.~Klingebiel, S.~A. Quintero,
  J.~Regenstein, N.~A. Browning, {R-ADVANCE: Rapid Adaptive Prediction for
  Vision-based Autonomous Navigation, Control, and Evasion}, WOL Journal of
  Field Robotics 35~(1) (2018) 91--100.
\newblock \href {https://doi.org/10.1002/rob.21744}
  {\path{doi:10.1002/rob.21744}}.

\bibitem{reynolds1987flocks}
C.~W. Reynolds, {Flocks, herds and schools: A distributed behavioral model},
  in: Proceedings of the 14th annual conference on Computer graphics and
  interactive techniques, 1987, pp. 25--34.

\bibitem{olfati2006flocking}
R.~Olfati-Saber, {Flocking for multi-agent dynamic systems: Algorithms and
  theory}, IEEE Transactions on automatic control 51~(3) (2006) 401--420.

\bibitem{la2010flocking}
H.~M. La, W.~Sheng, {Flocking control of multiple agents in noisy
  environments}, in: IEEE International Conference on Robotics and Automation,
  2010, pp. 4964--4969.

\bibitem{jia2017three}
Y.~Jia, J.~Du, W.~Zhang, L.~Wang, {Three-dimensional leaderless flocking
  control of large-scale small unmanned aerial vehicles}, Elsevier
  IFAC-PapersOnLine 50~(1) (2017) 6208--6213.

\bibitem{su2009flocking}
H.~Su, X.~Wang, Z.~Lin, {Flocking of multi-agents with a virtual leader}, IEEE
  transactions on automatic control 54~(2) (2009) 293--307.

\bibitem{quintero2013flocking}
S.~A. Quintero, G.~E. Collins, J.~P. Hespanha, {Flocking with fixed-wing UAVs
  for distributed sensing: A stochastic optimal control approach}, in: IEEE
  American Control Conference, 2013, pp. 2025--2031.

\bibitem{hung2016q}
S.-M. Hung, S.~N. Givigi, {A Q-learning approach to flocking with UAVs in a
  stochastic environment}, IEEE transactions on cybernetics 47~(1) (2016)
  186--197.

\bibitem{morihiro2007reinforcement}
K.~Morihiro, T.~Isokawa, H.~Nishimura, M.~Tomimasu, N.~Kamiura, N.~Matsui,
  {Reinforcement Learning Scheme for Flocking Behavior Emergence}, Journal of
  Advanced Computational Intelligence and Intelligent Informatics 11~(2) (2007)
  155--161.

\bibitem{xu2018multi}
Z.~Xu, Y.~Lyu, Q.~Pan, J.~Hu, C.~Zhao, S.~Liu, {Multi-vehicle flocking control
  with deep deterministic policy gradient method}, in: IEEE International
  Conference on Control and Automation, 2018, pp. 306--311.

\bibitem{ros2021online}
O.~Robotics, {ROS Home Page}, \url{https://www.ros.org/}, (Accessed:
  01.11.2021) (2021).

\bibitem{airsim2021online}
M.~Research, {Microsoft AirSim Home Page},
  \url{https://microsoft.github.io/AirSim/}, (Accessed: 01.11.2021) (2021).

\bibitem{gazebo2021online}
O.~S.~R. Foundation, {Gazebo Home Page}, \url{https://gazebosim.org/},
  (Accessed: 01.11.2021) (2021).

\bibitem{unrealengine2021online}
E.~Games, {Epic Games Unreal Engine Home Page},
  \url{https://www.unrealengine.com}, (Accessed: 01.11.2021) (2021).

\bibitem{watkins1992q}
C.~J. Watkins, P.~Dayan, {Q-learning}, Springer Machine learning 8~(3-4) (1992)
  279--292.
\newblock \href {https://doi.org/10.1007/978-1-4615-3618-5_4}
  {\path{doi:10.1007/978-1-4615-3618-5_4}}.

\bibitem{fotouhi2021deep}
A.~Fotouhi, M.~Ding, M.~Hassan, {Deep Q-Learning for Two-Hop Communications of
  Drone Base Stations}, MDPI Sensors 21~(6) (2021).
\newblock \href {https://doi.org/10.3390/s21061960}
  {\path{doi:10.3390/s21061960}}.

\bibitem{rummery1994line}
G.~A. Rummery, M.~Niranjan, {On-line Q-learning using connectionist systems},
  Vol.~37, University of Cambridge, 1994.

\bibitem{mnih2013playing}
V.~Mnih, K.~Kavukcuoglu, D.~Silver, A.~Graves, I.~Antonoglou, D.~Wierstra,
  M.~Riedmiller, {Playing Atari with Deep Reinforcement Learning},
  arXiv:1312.5602 (2013).

\bibitem{huang2019deep}
H.~Huang, Y.~Yang, H.~Wang, Z.~Ding, H.~Sari, F.~Adachi, {Deep reinforcement
  learning for UAV navigation through massive MIMO technique}, IEEE
  Transactions on Vehicular Technology 69~(1) (2019) 1117--1121.

\bibitem{VanHasselt2016}
H.~{Van Hasselt}, A.~Guez, D.~Silver, {Deep reinforcement learning with double
  Q-Learning}, AAAI Conference on Artificial Intelligence (2016) 2094--2100.

\bibitem{Wang2016}
Z.~Wang, T.~Schaul, M.~Hessel, H.~Hasselt, M.~Lanctot, N.~Freitas, {Dueling
  Network Architectures for Deep Reinforcement Learning}, in: PMLR
  International Conference on Machine Learning, Vol.~48, 2016, pp. 1995--2003.

\bibitem{hausknecht2015deep}
M.~Hausknecht, P.~Stone, {Deep Recurrent Q-learning for partially observable
  MDPS}, arXiv:1507.06527 (2015).

\bibitem{fortunato2017noisy}
M.~Fortunato, M.~G. Azar, B.~Piot, J.~Menick, I.~Osband, A.~Graves, V.~Mnih,
  R.~Munos, D.~Hassabis, O.~Pietquin, et~al., {Noisy networks for exploration},
  arXiv:1706.10295 (2017).

\bibitem{bellemare2017distributional}
M.~G. Bellemare, W.~Dabney, R.~Munos, {A distributional perspective on
  reinforcement learning}, in: PMLR International Conference on Machine
  Learning, 2017, pp. 449--458.

\bibitem{dabney2018distributional}
W.~Dabney, M.~Rowland, M.~G. Bellemare, R.~Munos, {Distributional reinforcement
  learning with quantile regression}, AAAI Conference on Artificial
  Intelligence (2018).

\bibitem{dabney2018implicit}
W.~Dabney, G.~Ostrovski, D.~Silver, R.~Munos, {Implicit quantile networks for
  distributional reinforcement learning}, in: PMLR International conference on
  machine learning, 2018, pp. 1096--1105.

\bibitem{hessel2018rainbow}
M.~Hessel, J.~Modayil, H.~Van~Hasselt, T.~Schaul, G.~Ostrovski, W.~Dabney,
  D.~Horgan, B.~Piot, M.~Azar, D.~Silver, {Rainbow: Combining improvements in
  deep reinforcement learning}, AAAI Conference on Artificial Intelligence
  (2018).

\bibitem{yang2019fully}
D.~Yang, L.~Zhao, Z.~Lin, T.~Qin, J.~Bian, T.-Y. Liu, {Fully parameterized
  quantile function for distributional reinforcement learning}, Advances in
  Neural Information Processing Systems 32 (2019) 6193--6202.

\bibitem{kapturowski2018recurrent}
S.~Kapturowski, G.~Ostrovski, J.~Quan, R.~Munos, W.~Dabney, {Recurrent
  experience replay in distributed reinforcement learning}, International
  Conference on Learning Representations (2018).

\bibitem{horgan2018distributed}
D.~Horgan, J.~Quan, D.~Budden, G.~Barth-Maron, M.~Hessel, H.~Van~Hasselt,
  D.~Silver, {Distributed prioritized experience replay}, arXiv:1803.00933
  (2018).

\bibitem{badia2020never}
A.~P. Badia, P.~Sprechmann, A.~Vitvitskyi, D.~Guo, B.~Piot, S.~Kapturowski,
  O.~Tieleman, M.~Arjovsky, A.~Pritzel, A.~Bolt, et~al., {Never give up:
  Learning directed exploration strategies}, arXiv:2002.06038 (2020).

\bibitem{badia2020agent57}
A.~P. Badia, B.~Piot, S.~Kapturowski, P.~Sprechmann, A.~Vitvitskyi, Z.~D. Guo,
  C.~Blundell, {Agent57: Outperforming the atari human benchmark}, in: PMLR
  International Conference on Machine Learning, 2020, pp. 507--517.

\bibitem{zhao2016deep}
D.~Zhao, H.~Wang, K.~Shao, Y.~Zhu, {Deep reinforcement learning with experience
  replay based on SARSA}, in: IEEE Symposium Series on Computational
  Intelligence, 2016, pp. 1--6.

\bibitem{williams1992simple}
R.~J. Williams, {Simple statistical gradient-following algorithms for
  connectionist reinforcement learning}, Springer Machine learning 8~(3-4)
  (1992) 229--256.

\bibitem{schulman2015trust}
J.~Schulman, S.~Levine, P.~Abbeel, M.~Jordan, P.~Moritz, {Trust Region Policy
  Optimization}, in: PMLR International Conference on Machine Learning,
  Vol.~37, 2015, pp. 1889--1897.

\bibitem{schulman2017proximal}
J.~Schulman, F.~Wolski, P.~Dhariwal, A.~Radford, O.~Klimov, {Proximal Policy
  Optimization Algorithms}, arXiv:1707.06347 (2017).

\bibitem{cobbe2020phasic}
K.~Cobbe, J.~Hilton, O.~Klimov, J.~Schulman, {Phasic Policy Gradient},
  arXiv:2009.04416 (2020).

\bibitem{liu2017stein}
Y.~Liu, P.~Ramachandran, Q.~Liu, J.~Peng, {Stein Variational Policy Gradient},
  arXiv:1704.02399 (2017).

\bibitem{lee2019stochastic}
A.~X. Lee, A.~Nagabandi, P.~Abbeel, S.~Levine, {Stochastic latent actor-critic:
  Deep reinforcement learning with a latent variable model}, arXiv:1907.00953
  (2019).

\bibitem{zhang2019ace}
S.~Zhang, H.~Yao, {ACE: An Actor Ensemble Algorithm for continuous control with
  tree search}, in: AAAI Conference on Artificial Intelligence, Vol.~33, 2019,
  pp. 5789--5796.
\newblock \href {https://doi.org/10.1609/aaai.v33i01.33015789}
  {\path{doi:10.1609/aaai.v33i01.33015789}}.

\bibitem{zhang2019dac}
S.~Zhang, S.~Whiteson, {DAC: The double actor-critic architecture for learning
  options}, arXiv:1904.12691 (2019).

\bibitem{heess2015memory}
N.~Heess, J.~J. Hunt, T.~P. Lillicrap, D.~Silver, {Memory-Based Control with
  Recurrent Neural Networks}, arXiv:1512.04455 (2015).

\bibitem{lillicrap2015continuous}
T.~P. Lillicrap, J.~J. Hunt, A.~Pritzel, N.~Heess, T.~Erez, Y.~Tassa,
  D.~Silver, D.~Wierstra, {Continuous control with deep reinforcement
  learning}, arXiv:1509.02971 (2015).

\bibitem{fujimoto2018addressing}
S.~Fujimoto, H.~Hoof, D.~Meger, {Addressing function approximation error in
  actor-critic methods}, in: PMLR International Conference on Machine Learning,
  2018, pp. 1587--1596.

\bibitem{haarnoja2018soft}
T.~Haarnoja, A.~Zhou, P.~Abbeel, S.~Levine, {Soft Actor-Critic: Off-policy
  maximum entropy deep reinforcement learning with a stochastic actor}, in:
  PMLR International Conference on Machine Learning, 2018, pp. 1861--1870.

\bibitem{barth2018distributed}
G.~Barth-Maron, M.~W. Hoffman, D.~Budden, W.~Dabney, D.~Horgan, D.~Tb,
  A.~Muldal, N.~Heess, T.~Lillicrap, {Distributed distributional deterministic
  policy gradients}, arXiv:1804.08617 (2018).

\bibitem{mnih2016asynchronous}
V.~Mnih, A.~P. Badia, M.~Mirza, A.~Graves, T.~Lillicrap, T.~Harley, D.~Silver,
  K.~Kavukcuoglu, {Asynchronous methods for deep reinforcement learning}, in:
  PMLR International Conference on Machine Learning, 2016, pp. 1928--1937.

\bibitem{heess1707emergence}
N.~Heess, D.~TB, S.~Sriram, J.~Lemmon, J.~Merel, G.~Wayne, Y.~Tassa, T.~Erez,
  Z.~Wang, S.~Eslami, et~al., {Emergence of locomotion behaviours in rich
  environments}, arXiv:1707.02286 (2017).

\bibitem{alfredo2017efficient}
C.~Alfredo, C.~Humberto, C.~Arjun, {Efficient parallel methods for deep
  reinforcement learning}, in: The Multi-disciplinary Conference on
  Reinforcement Learning and Decision Making, 2017, pp. 1--6.

\bibitem{wang2016sample}
Z.~Wang, V.~Bapst, N.~Heess, V.~Mnih, R.~Munos, K.~Kavukcuoglu, N.~de~Freitas,
  {Sample efficient actor-critic with experience replay}, arXiv:1611.01224
  (2016).

\bibitem{gruslys2017reactor}
A.~Gruslys, W.~Dabney, M.~G. Azar, B.~Piot, M.~Bellemare, R.~Munos, {The
  reactor: A fast and sample-efficient actor-critic agent for reinforcement
  learning}, arXiv:1704.04651 (2017).

\bibitem{wu2017scalable}
Y.~Wu, E.~Mansimov, R.~B. Grosse, S.~Liao, J.~Ba, {Scalable Trust-Region method
  for deep reinforcement learning using Kronecker-Factored Approximation}, in:
  Advances in Neural Information Processing Systems, 2017, pp. 5279--5288.

\bibitem{lowe2017multi}
R.~Lowe, Y.~Wu, A.~Tamar, J.~Harb, P.~Abbeel, I.~Mordatch, {Multi-Agent
  Actor-Critic for mixed cooperative-competitive environments},
  arXiv:1706.02275 (2017).

\bibitem{ackermann2019reducing}
J.~Ackermann, V.~Gabler, T.~Osa, M.~Sugiyama, {Reducing overestimation bias in
  multi-agent domains using double centralized critics}, arXiv:1910.01465
  (2019).

\bibitem{iqbal2019actor}
S.~Iqbal, F.~Sha, {Actor-Attention-Critic for Multi-Agent Reinforcement
  Learning}, in: PMLR International Conference on Machine Learning, 2019, pp.
  2961--2970.

\bibitem{espeholt2018impala}
L.~Espeholt, H.~Soyer, R.~Munos, K.~Simonyan, V.~Mnih, T.~Ward, Y.~Doron,
  V.~Firoiu, T.~Harley, I.~Dunning, et~al., {IMPALA: Scalable distributed
  deep-rl with importance weighted actor-learner architectures}, in: PMLR
  International Conference on Machine Learning, 2018, pp. 1407--1416.

\bibitem{espeholt2019seed}
L.~Espeholt, R.~Marinier, P.~Stanczyk, K.~Wang, M.~Michalski, {Seed RL:
  Scalable and Efficient Deep-RL with accelerated central inference},
  arXiv:1910.06591 (2019).

\bibitem{VanHasselt2010}
H.~Hasselt, {Double Q-learning}, Advances in Neural Information Processing
  Systems 23 (2010) 2613--2621.

\bibitem{burda2018exploration}
Y.~Burda, H.~Edwards, A.~Storkey, O.~Klimov, {Exploration by random network
  distillation}, arXiv:1810.12894 (2018).

\bibitem{sutton2000policy}
R.~S. Sutton, D.~A. McAllester, S.~P. Singh, Y.~Mansour, {Policy gradient
  methods for reinforcement learning with function approximation}, in: Advances
  in Neural Information Processing Systems, 2000, pp. 1057--1063.

\bibitem{shin2019obstacle}
S.-Y. Shin, Y.-W. Kang, Y.-G. Kim, {Obstacle Avoidance Drone by Deep
  Reinforcement Learning and Its Racing with Human Pilot}, MDPI Applied
  Sciences 9~(24) (2019).
\newblock \href {https://doi.org/10.3390/app9245571}
  {\path{doi:10.3390/app9245571}}.

\bibitem{liu2016stein}
Q.~Liu, D.~Wang, {Stein variational gradient descent: A general purpose
  bayesian inference algorithm}, arXiv:1608.04471 (2016).

\bibitem{wierstra2010recurrent}
D.~Wierstra, A.~F{\"o}rster, J.~Peters, J.~Schmidhuber, {Recurrent Policy
  Gradients}, Logic Journal of the IGPL 18~(5) (2010) 620--634.

\bibitem{martens2015optimizing}
J.~Martens, R.~Grosse, {Optimizing Neural Networks with Kronecker-Factored
  Approximate Curvature}, in: PMLR International conference on machine
  learning, 2015, pp. 2408--2417.

\bibitem{munos2016safe}
R.~Munos, T.~Stepleton, A.~Harutyunyan, M.~G. Bellemare, {Safe and efficient
  off-policy reinforcement learning}, arXiv:1606.02647 (2016).

\bibitem{andrychowicz2020matters}
M.~Andrychowicz, A.~Raichuk, P.~Stanczyk, M.~Orsini, S.~Girgin, R.~Marinier,
  L.~Hussenot, M.~Geist, O.~Pietquin, M.~Michalski, et~al., {What Matters In
  On-Policy Reinforcement Learning? A Large-Scale Empirical Study}, CoRR
  abs/2006.05990 (2020).

\end{thebibliography}


\bio{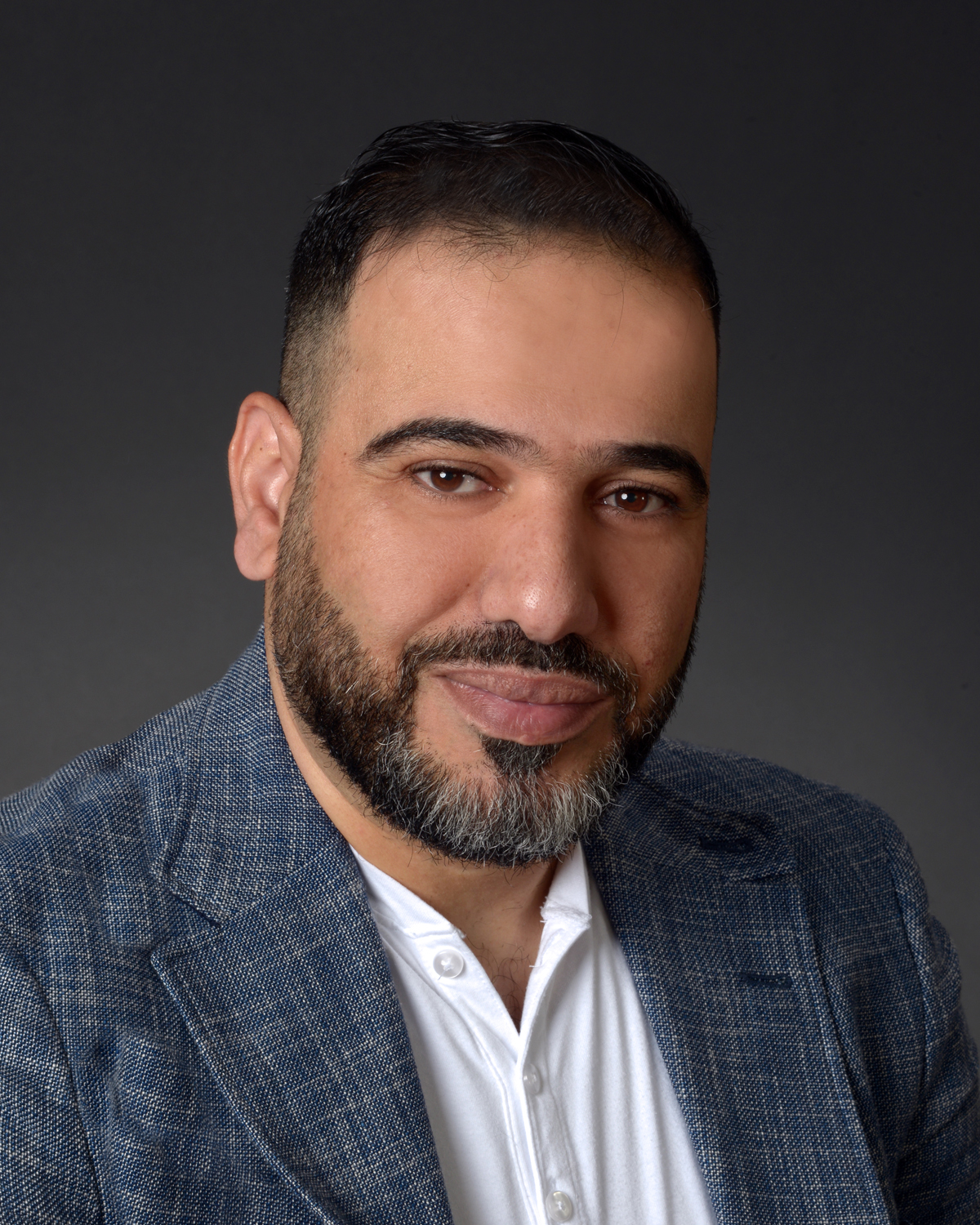}{\textbf{Fadi AlMahamid}}
received the B.Sc. degree (Hons.) in Computer Science from Princess Sumaya University for Technology (PSUT), Amman, Jordan, in 2001, and M.Sc. degree (Hons.) in Computer Science from New York Institute of Technology (NYIT), Amman, Jordan, in 2003. Also, he obtained another M.Sc. from the University of Western Ontario (UWO), London, Ontario, Canada, in 2019, where he is currently pursuing the Ph.D. degree in Software Engineering with the Department of Electrical and Computer Engineering. He has extensive industry experience of more than 15 years. His current research interests include machine learning, autonomous vehicles focusing on navigation problems, IoT architectures, and sensor data analytics. 
\endbio

\bio{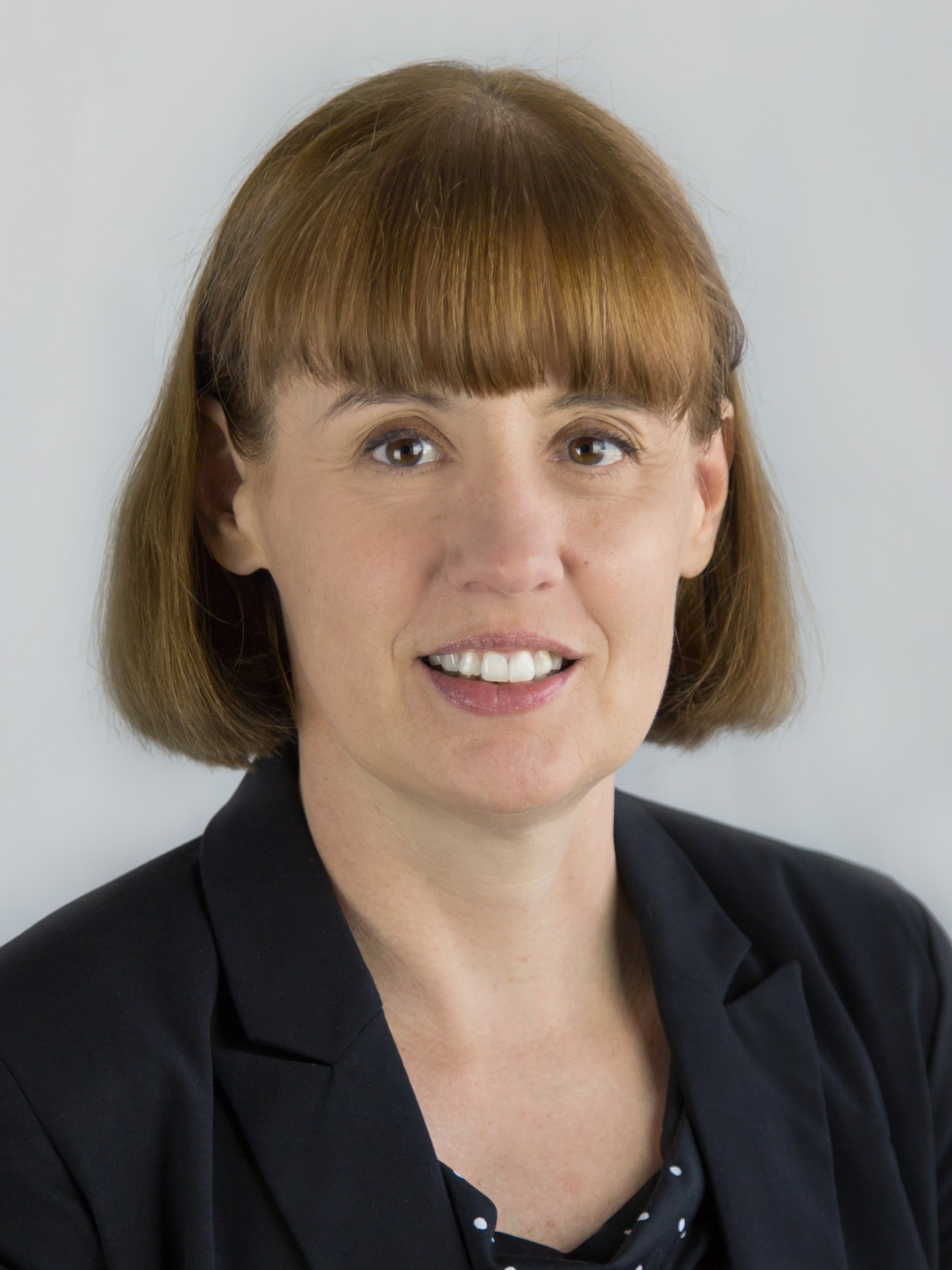}{\textbf{Katarina Grolinger}}
received the B.Sc. and M.Sc. degrees in mechanical engineering from the University of Zagreb, Croatia, and the M.Eng. and Ph.D. degrees in software engineering from Western University, London, Canada. She is currently an Assistant Professor with the Department of Electrical and Computer Engineering, Western University. She has been involved in the software engineering area in academia and industry, for over 20 years. Her current research interests include machine learning, sensor data analytics, data management, and the IoT. 
\endbio

\end{document}